\documentclass[a4paper,11pt]{article}

\usepackage{microtype}
\usepackage{graphicx}
\usepackage{booktabs} %
\usepackage{tabularx}
\usepackage{notation}
\usepackage[dvipsnames]{xcolor}
\usepackage{hyperref}
\usepackage{mathtools}
\usepackage[export]{adjustbox}
\DeclarePairedDelimiter{\ceil}{\lceil}{\rceil}
\DeclarePairedDelimiter\floor{\lfloor}{\rfloor}
\usepackage{algorithm}
\usepackage[noend]{algorithmic}

\usepackage{stmaryrd}
\usepackage{fullpage}
\usepackage{authblk}
\usepackage{natbib}

\usepackage{amsthm}
\usepackage{amsmath}
\usepackage{amsfonts}
\usepackage{amssymb}
\usepackage[capitalise,nameinlink]{cleveref}
\usepackage[shortlabels]{enumitem}

\usepackage{multirow}
\usepackage{subcaption}

\crefname{section}{Sec.}{Sec.}
\crefname{appendix}{Sec.}{Sec.}
\crefname{theorem}{Thm.}{Thm.}
\crefname{corollary}{Cor.}{Cor.}
\crefname{proposition}{Prop.}{Prop.}
\crefname{definition}{Def.}{Def.}
\crefname{table}{Tab.}{Tab.}
\crefname{algorithm}{Alg.}{Alg.}
\crefname{example}{Ex.}{Ex.}
\crefformat{footnote}{#2\footnotemark[#1]#3}

\newtheorem{theorem}{Theorem}[section]
\newtheorem{corollary}{Corollary}[theorem]
\newtheorem{proposition}{Proposition}[section]
\newtheorem{lemma}[theorem]{Lemma}

\newtheorem{example}[theorem]{Example}

\theoremstyle{definition}
\newtheorem{definition}{Definition}[section]

\newcommand{\abs}[1]{\left\lvert#1\right\rvert}

\newcommand{\proj}[2]{\left.{#1}\right\rvert_{#2}}

\newcommand{\R}{\ensuremath{\mathbb{R}}}
\newcommand{\N}{\ensuremath{\mathbb{N}}}

\newcommand{\bigO}{\ensuremath{\mathcal{O}}}

\newcommand{\node}{{n}}



\newcommand{\inputUnit}{\textsc{Input}}
\newcommand{\sumUnit}{\textsc{Sum}}
\newcommand{\prodUnit}{\textsc{Product}}
\newcommand{\suppCirc}{\textsc{Support}}

\newcommand{\LineIf}[2]{     
    \STATE \algorithmicif\ {#1}\ \algorithmicthen\ {#2} 
}
\newcommand{\LineIfElse}[3]{     
    \STATE \algorithmicif\ {#1}\ \algorithmicthen\ {#2}\ \algorithmicelse\ {#3}
}

\title{A Compositional Atlas of Tractable Circuit Operations:\\ From Simple Transformations to Complex Information-Theoretic Queries}

 \author{Antonio Vergari$^{1}$}
 \author{YooJung Choi$^{1}$}
 \author{Anji Liu$^{1}$}
 \author{Stefano Teso$^{2}$}
 \author{Guy Van den Broeck$^{1}$\vspace{5pt}}

\affil{%
  {$^{1}$Computer Science Department, University of California, Los Angeles, USA \hfill\texttt{\{aver|yjchoi|liuanji|guyvdb\}@cs.ucla.edu}} \\
  {$^{2}$Department of Computer Science, University of Trento, Italy \hfill \\ \texttt{stefano.teso@unitn.it}}
}

\begin{document}

\maketitle
\begin{abstract}
Circuit representations are becoming the lingua franca to express and reason about tractable generative and discriminative models.
In this paper, we show how complex inference scenarios for these models that commonly arise in machine learning---from computing the expectations of decision tree ensembles to information-theoretic divergences of deep mixture models---can be represented in terms of tractable modular operations over circuits.
Specifically, we characterize the tractability of a vocabulary of simple transformations---sums, products, quotients, powers, logarithms, and exponentials---in terms of sufficient structural constraints of the circuits they  operate on, and present novel hardness results for the cases in which these properties are not satisfied.
Building on these operations, we derive a unified framework for reasoning about tractable models that  generalizes several results in the literature and opens up novel tractable inference scenarios.
\end{abstract}

\section{Introduction}
\label{sec:intro}

Many core computational tasks in machine learning (ML) and AI involve solving complex \textit{integrals}, such as expectations, that often turn out to be intractable.
A fundamental question then arises: \textit{under which conditions do these quantities admit tractable computation?} That is, when can we compute them efficiently without resorting to approximations or heuristics?
Consider for instance the Kullback-Leibler divergence (KLD) between two distributions $p$ and $q$: $\mathbb{D}_{\mathsf{KL}}(\p\parallel\q)=\int \p(\x)\log(\p(\x)/\q(\x))d\X$.
Characterizing its tractability can have important applications in learning, 
approximate inference~\citep{shih2020probabilistic}, and model compression~\citep{LiangXAI17}.

This ``quest'' for tracing the tractability of certain quantities of  
interest---henceforth called \textit{queries}---has been carried out several times, often independently, for different model classes in ML and AI, and crucially for each query in isolation.
Here, we take a different path and introduce a general framework under which the tractability of many complex queries can be traced in a unified manner. 

To do so, we focus on circuit representations~\citep{choi2020pc} that guarantee exact computation of integrals of interest if the circuit satisfies specific structural properties.
They subsume many generative models---probabilistic circuits such as Chow-Liu trees~\citep{chow1968approximating}, hidden Markov models (HMMs)~\citep{rabiner1986introduction}, sum-product networks (SPNs)~\citep{poon2011sum}, and other deep mixture models---as well as discriminative ones---including 
decision trees~\citep{khosravi2020handling,correia2020joints} and deep regressors~\citep{khosravi2019tractable}---thus enabling a unified treatment of many inference scenarios.

\begin{figure}
    \centering
    \includegraphics[width=.5\columnwidth]{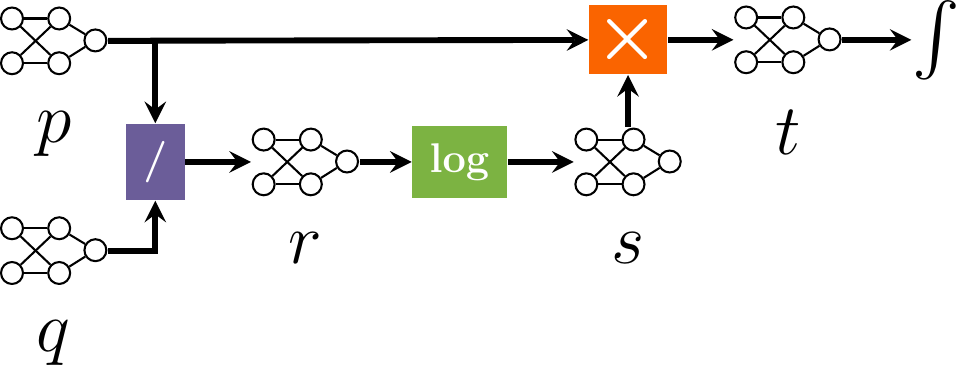}\hfill
        \includegraphics[width=.35\columnwidth]{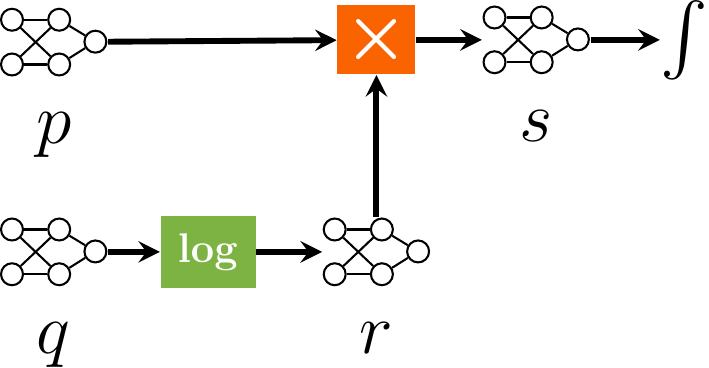}
    \caption{Computational pipelines of the KLD (left) and cross entropy (right) over two distributions $p$ and $q$ encoded as circuits, where the intermediate computations ($r$, $s$ and $t$) are represented as circuits as well.
    }
    \label{fig:pipelines}
\end{figure}

We represent complex queries as computational pipelines whose intermediate operations transform and combine the input circuits into other circuits.
This representation enables us to analyze tractability by ``propagating'' the sufficient conditions through all intermediate steps.
For instance, consider the pipeline for computing the KLD of $\p$ and $\q$, two distributions represented by circuits, as shown in \cref{fig:pipelines}. By tracing the tractability conditions of the quotient, logarithm, and product over circuits such that the output circuit (i.e., $t$) admits tractable integration, we can derive a set of minimal sufficient conditions for the input circuits.
That is, we can identify a general class of models that supports tractable computation of the KLD.

By re-using the tractability conditions of these simple operations and their algorithms as sub-routines across queries, we are able to compositionally answer many other complex queries.
For instance, we can reuse the logarithm and product operations in the KLD pipeline to reason about the tractability of cross entropy (\cref{fig:pipelines}).
These sub-routines can be easily implemented in circuit libraries such as Juice~\citep{dang2021juice} or SPFlow~\citep{molina2019spflow}.

We make the following contributions. 
(1) We provide a grammar of simple circuit transformations---products, quotients, powers, logarithms, and exponentials---and
introduce sufficient conditions for their tractability while proving their hardness otherwise (\cref{tab:tract-op}); 
(2) Using this grammar, we unify inference algorithms proposed in the literature for specific representations and extend their scope towards larger model classes; 
(3) We provide novel tractability and hardness results of complex information-theoretic queries including several widely used entropies and divergences~(\cref{tab:queries}).

\section{Circuit Representations}

Circuits represent functions as parameterized computational graphs. 
By imposing certain structural constraints on these graphs, or verifying their presence, we can guarantee the tractability of certain operations over the encoded functions.
As such, circuits provide \textit{a language for building and reasoning about tractable representations}.
We proceed by introducing the basic rules of this language.

We denote random variables by uppercase letters ($X$) and their values/assignments by lowercase ones ($x$). 
Sets of variables and their assignments are denoted by bold uppercase ($\X$) and bold lowercase ($\x$) letters, respectively.

\begin{definition}[Circuit]
\label{def:circuits}
A circuit $\p$ over variables $\X$ is a parameterized computational graph encoding a function $\p(\X)$
and comprising three kinds of computational units:
\textit{input}, \textit{product}, and \textit{sum}. 
Each inner unit $n$ (i.e., product or sum unit) receives inputs from some other units, denoted $\ch(n)$.
Each unit $n$ encodes a function $\p_{n}$ as follows:
\begin{equation*}
{\p}_n(\X)=
\begin{cases}
    l_n({\phi(n)}) &\text{if $n$ is an input unit} \\
    \prod_{c\in\ch(n)} \p_c(\X) &\text{if $n$ is a product unit} \\
    \sum_{c\in\ch(n)} \theta_{c} \p_c(\X) &\text{if $n$ is a sum unit}
\end{cases}
\label{eq:EVI}
\end{equation*}
where $\theta_{c}\in\R$ are the sum parameters,
and input units encode parameterized functions $l_n$ over variables $\phi(n) \subseteq \X$, also called their \textit{scope}.
The scope of an inner unit
is the union of the scopes of its inputs:  $\phi(n)=\bigcup_{c \in \ch(n)}\phi(c)$.
The output unit of the circuit is the last unit (i.e., with out-degree 0) in the graph, encoding $p(\X)$.
The \emph{support} of $\p$ is the set of all complete states for $\X$ for which the output of $\p$ is non-zero: $\supp(\p)=\{\x\in\val(\X)\,|\, \p(\x)\neq0\}$.
\end{definition}

\begin{figure}[t]
    \centering
    \begin{subfigure}[t]{0.39\columnwidth}
        \centering
        \includegraphics[width=.99\columnwidth]{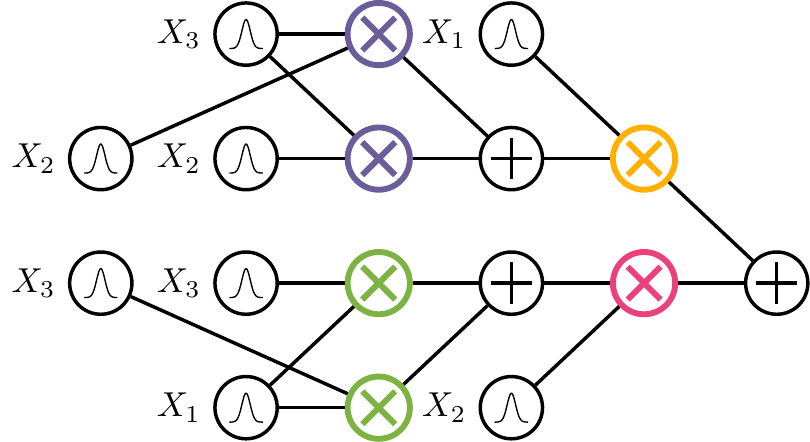}
        \caption{Decomposable\label{fig:dec-circuit}}
    \end{subfigure}\hspace{30pt}
    \begin{subfigure}[t]{0.39\columnwidth}
        \centering
        \includegraphics[width=.99\columnwidth]{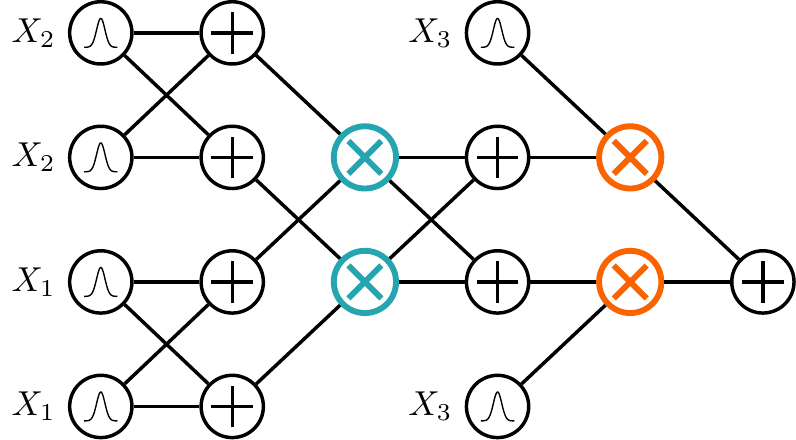}
    \caption{Structured-decomposable\label{fig:sd-circuit}}
    \end{subfigure}\\[30pt]
    \begin{subfigure}[]{0.99\columnwidth}
        \centering
        \begin{minipage}{.125\columnwidth}
            \centering
            \includegraphics[width=.99\columnwidth]{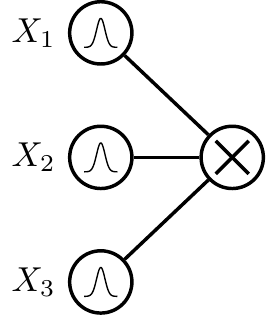}
        \end{minipage}
        \begin{minipage}{.35\columnwidth}
            \centering
            \includegraphics[width=.9\columnwidth]{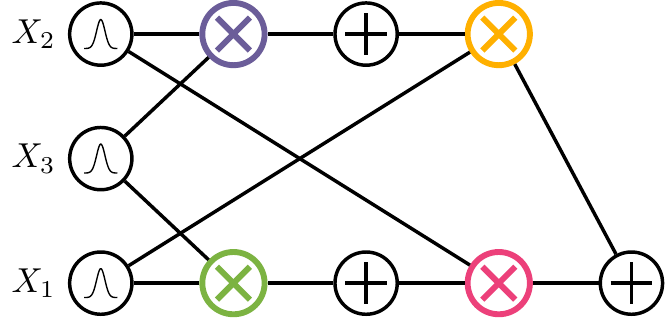}
        \end{minipage}
        \begin{minipage}{.35\columnwidth}
            \centering
            \includegraphics[width=.9\columnwidth]{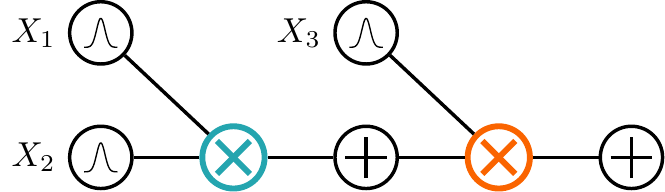}
        \end{minipage}
    \caption{Omni-compatible circuits for $p(\X)=p(X_1)p(X_2)p(X_3)$}\label{fig:omni-circuit}
    \end{subfigure}
    \caption{\textit{Examples of circuit representations} with different structural properties. The feedforward order is from left to right; input units are labeled by their scope; and sum parameters are omitted for visual clarity.
    Product units of the rearranged omni-compatible factorization (bottom) are color-coded with those of matching scope in the top circuits.
    }
    \label{fig:circuits}
\end{figure}

Circuits can be understood as compact representations of polynomials with possibly an exponential number of terms, whose indeterminates are the functions encoded by the input units. 
These functions are assumed to be simple enough to allow tractable computations of the operations discussed in this paper.
\cref{fig:circuits} shows some examples of circuits.

A \textit{probabilistic circuit} (PC)~\citep{choi2020pc} represents a (possibly unnormalized) probability distribution by encoding its probability mass, density, or a combination thereof.
\begin{definition}[Probabilistic circuit]
\label{def:pcs}
A PC over variables $\X$ is a circuit encoding a function $\p$ that is non-negative for all values of $\X$; \ie $\forall \x\in\val(\X):p(\x)\geq 0$.
\end{definition}

From here on, we will assume a PC to have positive sum parameters
and input units that model valid (unnormalized) distributions, 
which is a sufficient condition to satisfy the above definition.
Moreover, w.l.o.g.\ we will assume that each layer of a circuit alternates between sum and product units and that every product unit $n$ receives only two inputs, \ie $\p_n(\X)=\p_{c_1}(\X)\p_{c_2}(\X)$.
These conditions can easily be enforced on any circuit in exchange for a polynomial increase in its size~\citep{vergari2015simplifying,peharz2020einsum}.

Computing (functions of) $\p(\X)$, or in other words performing \textit{inference}, can be done by evaluating its computational graph.
Hence, the computational cost of inference on a circuit
is a function of its \textit{size}, defined as the number of edges and denoted as $|\p|$.
For instance, querying the value of $\p$ for a complete assignment $\x$ equals its \textit{feedforward} evaluation---inputs before outputs---and therefore is linear in $|\p|$.
Other common inference scenarios such as function integration---which translate to \textit{marginal inference} in the context of probability distributions---can be tackled in linear time with circuits that exhibit certain structural properties, as discussed next.

\subsection{Structural Properties of Circuits}
\label{sec:struct-prop}

Structural constraints on the computational graph of a circuit \wrt its scope or support can provide sufficient and/or necessary conditions for certain queries to be computed exactly in polytime.
Therefore, one can characterize inference scenarios, also known as classes of queries, in terms of the structural properties realizing these constraints.
Moreover, these constraints help understand how circuits generalize several classical tractable model classes, such as mixture models, bounded-treewidth probabilistic graphical models (PGMs), decision trees, and compact logical function representations.
It follows that all our results in the following sections automatically translate to these model classes. %
We now define the structural properties that this work will focus on, referring to~\citet{choi2020pc} for more details.

\begin{definition}[Smoothness]
A circuit is \textit{smooth} if for every sum unit $n$, its inputs depend on the same variables: $\forall\, c_1, c_2\in\ch(n), \phi(c_1)=\phi(c_2)$.
\end{definition}
Smooth PCs generalize homogeneous and shallow mixture models~\citep{mclachlan2019finite} to deep and hierarchical models.
For instance, a Gaussian mixture model (GMM) can be represented as a smooth PC with a single sum unit over as many input units as mixture components, each encoding a (multivariate) Gaussian density. 

\begin{definition}[Decomposability]
\label{def:dec}
A circuit is \textit{decomposable} 
if the inputs of every product unit $n$ depend on disjoint sets of variables: $\ch(n)=\{c_1,c_2\}, \phi(c_1)\cap\phi(c_2)=\emptyset$.
\end{definition}
Decomposable product units encode local factorizations.
That is, a decomposable product unit $n$ over variables $\X$ encodes $\p_{n}(\X)=\p_{1}(\X_1)\cdot\p_{2}(\X_2)$ 
where $\X_1$ and $\X_2$ form a partition of $\X$.
Taken together, decomposability and smoothness are a sufficient and necessary condition for performing tractable integration over arbitrary sets of variables in a single feedforward pass, as they enable larger integrals to be efficiently decomposed into smaller ones~\citep{choi2020pc}.

\begin{proposition}[Tractable integration]
\label{prop:trac-integration}
Let $\p$ be a smooth and decomposable circuit over $\X$ with input functions that can be tractably integrated. Then the 
integral $\int_{\z\in\val(\Z)}p(\y,\z)d\Z$
can be computed exactly in $\Theta(\abs{\p})$ time for any $\Y\subseteq\X,\y\in\val(\Y),\Z\!=\!\X\setminus\Y$.
\end{proposition}

Many complex queries involve integration as the last step.  It is therefore convenient that any intermediate operations preserve at least decomposability;  smoothness is less of an issue, as it can be enforced in polytime~\citep{shih2019smoothing}.
Smooth and decomposable PCs with millions of parameters can be efficiently learned from data~\citep{peharz2020einsum}.

A key additional constraint over scope decompositions is \textit{compatibility}.
Intuitively, two decomposable circuits are compatible if they can be rearranged in polynomial time\footnote{By changing the order in which n-ary product units are turned into a series of binary product units. 
} such that their respective product units, once matched by scope, decompose in the same way. We formalize this with the following inductive definition.

\begin{definition}[Compatibility]
    \label{def:compatibility}
    Two circuits $\p$ and $\q$ over variables $\X$ are \textit{compatible} if (1) they are smooth and decomposable and (2) any pair of product units $n\!\in\!\p$ and $m\!\in\!\q$ with the same scope can be rearranged into binary products that are mutually compatible and decompose in the same way:
    $(\scope(n)\!=\!\scope(m)) \!\implies\! (\scope(n_i)\!=\!\scope(m_i), \text{ $n_i$ and $m_i$ are compatible}) $ 
    for some rearrangement of the inputs of $n$ (resp.\ $m$) into $n_1,n_2$ (resp.\ $m_1,m_2$). 
\end{definition}
\begin{definition}[Structured-decomposability]
A circuit is \textit{structured-decomposable} if it is compatible with itself.
\end{definition}
\vspace{-3pt}
Not all decomposable circuits are structured-decomposable (see \cref{fig:dec-circuit,fig:sd-circuit}), but some can be rearranged to be compatible with any decomposable circuit (see \cref{fig:omni-circuit}).
\begin{definition}[Omni-compatibility]
A decomposable circuit $\p$ over $\X$ is \textit{omni-compatible} if it
is
compatible with any smooth and decomposable circuit over $\X$. 
\end{definition}
\vspace{-3pt}
For example, in \cref{fig:omni-circuit}, the fully factorized product unit $p(\X)=p_1(X_1)p_2(X_2)p_3(X_3)$ can be rearranged into $p_1(X_1)(p_2(X_2)p_3(X_3))$ and $p_2(X_2)(p_1(X_1)p_3(X_3))$ to match the yellow and pink products in 
\cref{fig:dec-circuit}.
We can easily see that omni-compatible 
circuits must assume the form of mixtures of fully-factorized models; \ie 
$\sum_{i}\theta_i\prod_{j}\p_{i,j}(X_j)$.
For example, an additive ensemble of decision trees over variables $\X$ can be represented as an omni-compatible circuit (cf. \cref{ex:decision-tree}).
Also note that if a circuit is compatible with a non-omni-compatible circuit, then it must be structured-decomposable.

\begin{definition}[Determinism]
\label{def:determinism}
A circuit is \textit{deterministic} if the inputs $\ch(n)$ of every sum unit $n$ have disjoint supports: $\forall\, c_1,c_2\in\ch(n), c_1\neq c_2 \implies \supp(c_1)\cap\supp(c_2)=\emptyset$.
\end{definition}

Analogously to decomposability,
{determinism} induces a recursive partitioning over the support of a circuit. 
For a deterministic sum unit $n$, the partitioning of its support can be made explicit by introducing an indicator function per each of its inputs, \ie $\sum_{c\in\ch(n)}\theta_{c} \p_c(\x) = \sum_{c\in\ch(n)}\theta_{c} \p_c(\x)\id{\x\in\supp(\p_{c})}$.

Determinism allows for tractable maximization of a circuit~\citep{choi2020pc}. 
While we are not investigating 
maximization in this work, determinism will still play a crucial role in the next sections.
Moreover, bounded-treewidth PGMs, such as Chow-Liu trees~\citep{chow1968approximating} and thin junction trees~\citep{bach2001thin}, can be represented as a smooth, deterministic, and decomposable PC via \textit{compilation}~\citep{darwiche2009modeling,dang2020strudel}.
Probabilistic sentential decision diagrams (PSDDs)~\citep{kisa2014probabilistic} are deterministic and structured-decomposable PCs 
that can be efficiently learned from data~\citep{dang2020strudel}.

\begin{table*}[t]
    \caption{
    \textbf{\textit{Tractability and hardness of simple circuit operations}}.
    Tractable conditions on inputs translate to conditions on outputs.
    E.g., consider the quotient of two circuits $p$ and $q$: if they are compatible (Cmp) and $q$ is deterministic (Det), then the output is decomposable (Dec), and deterministic if $p$ is also (+) deterministic and structured-decomposable (SD) if both $p$ and $q$ are.
    Hardness is for representing the result as a smooth (Sm) and decomposable circuit without some input condition.
    }
    \vspace{-0.5em}
    \label{tab:tract-op}
    \centering
    \scalebox{0.65}{
    \begin{tabular}{lclll@{\hspace{0.3em}}ll@{\hspace{0.3em}}l}
        \toprule
        \multicolumn{2}{c}{\multirow{2}{*}[-0.08cm]{Operation}} & \multicolumn{4}{c}{Tractability} & \multicolumn{2}{c}{\multirow{2}{*}[-0.08cm]{Hardness}} \\
        \cmidrule(lr){3-6}
        \multicolumn{2}{c}{} & {\small Input conditions} & {\small Output conditions} & \multicolumn{2}{l}{\small Time Complexity} & \multicolumn{1}{c}{}\\
        \midrule
        \textsc{Sum} & $\theta_1 \p + \theta_2 \q$ & (+Cmp)  & (+SD) & $\bigO(|\p|\!+\!|\q|)$ &  & NP-hard for Det out & \citep{shen2016tractable} \\
        \textsc{Product} & $\p \cdot \q$ & Cmp (+Det, +SD) & Dec (+Det, +SD) & $\bigO(|\p||\q|)$ & 
        (\cref{thm:prod-pcs}) & \#P-hard w/o Cmp & (\cref{thm:prod-dec}) \\ %
        \multirow{2}{*}{\textsc{Power}} & $\p^n,n\in\N$ & SD (+Det) & SD (+Det) & $\bigO(|\p|^n)$ & (\cref{thm:trac-natural-pow}) & \#P-hard w/o SD & (\cref{thm:hardness-natural-pow,thm:hardness-natural-pow-sd}) \\ 
              & $\p^\alpha,\alpha\in\R$ & Sm, Dec, Det (+SD)& Sm, Dec, Det (+SD) & $\bigO(|\p|)$ & (\cref{thm:real-pow-det}) & \#P-hard w/o Det & (\cref{thm:inv})\\
        \textsc{Quotient} & $\p/\q$ & Cmp; $\q$ Det (+$p$ Det,+SD) & Dec (+Det,+SD) 
        & $\bigO(|\p||\q|)$ & (\cref{thm:ratio-pcs}) & \#P-hard w/o Det & (\cref{thm:hardness-ratio-pcs}) \\
        \textsc{Log} & $\log(\p)$ & Sm, Dec, Det & Sm, Dec & $\bigO(|\p|)$ & (\cref{thm:log-det}) & \#P-hard w/o Det & (\cref{thm:hardness-log}) \\
        \textsc{Exp} & $\exp (p)$ & linear %
        & SD %
        & $\bigO(\abs{\p})$ & (\cref{thm:exp-linear}) & \#P-hard & (\cref{thm:hardness-exp}) \\
        \bottomrule
    \end{tabular}
}
\end{table*}

\section{From Simple Circuit Transformations\ldots}
\label{sec:ops}

This section aims to build and analyze an atlas of simple operations over circuits which can then be composed into more complex operations and queries.
Specifically, for each of these operations we are interested in characterizing (1) its tractability in terms of the structural properties of its input circuits, 
and (2) its closure \wrt these properties, i.e.\ whether they are preserved in the output circuit, while (3) tracing the hardness of representing the output as a decomposable circuit when some property is unmet.
Given limited space, we summarize all our main results in \cref{tab:tract-op} and 
prove the corresponding statements in the Appendix.
\begin{theorem}
The tractability and hardness results for simple circuit operations in \cref{tab:tract-op} hold.
\end{theorem}

\subsection{Sum of Circuits}
The simplest operations we can consider are sums and products: a natural choice given that our circuits comprise sum and product units. 
The operation of summing two circuits $\p(\Z)$ and $\q(\Y)$ is defined as $s(\X) = \theta_{1}\cdot\p(\Z) + \theta_{2}\cdot\q(\Y)$
for $\X=\Z\cup\Y$ and two real parameters $\theta_1,\theta_2\in\R$.
This operation, which is at the core of additive ensembles of tractable representations,\footnote{If $\p$ and $\q$ are PCs, $s$ realizes a monotonic mixture model if $\theta_1,\theta_2>0$ and $\theta_1+\theta_2=1$, which is clearly still a PC.} can be realized by introducing a single sum unit 
that takes as input $\p$ and $\q$. 
Summation applies to any input circuits, regardless of structural assumptions, and it preserves several properties.
In particular, if $\p$ and $\q$ are decomposable then $s$ is also decomposable; moreover, if they are compatible then $s$ is structured-decomposable as well as compatible with $\p$ and $\q$.
However, representing a sum as a deterministic circuit is known to be 
NP-hard~\citep{shen2016tractable}, even for compatible and deterministic inputs.

\subsection{Product of Circuits}

Multiplication is at the core of many of the compositional queries in \cref{sec:queries}.
The product of two circuits $\p(\Z)$ and $\q(\Y)$ can be expressed as $\m(\X) = \p(\Z)\cdot\q(\Y)$
for variables $\X=\Z\cup\Y$. 
If $\Z$ and $\Y$ are disjoint, the product $m$ is already decomposable.
\citet{shen2016tractable} proved that representing the product of two decomposable circuits as a decomposable circuit is NP-hard, even if they are deterministic. 
We prove in \cref{thm:prod-dec} that it is \#P-hard even for structured-decomposable and deterministic circuits.

Recently, \citet{shen2016tractable} introduced an efficient algorithm for the product of two 
compatible, deterministic PCs (namely PSDDs).
We prove that compatibility alone is sufficient for tractable product computation of any two circuits (\cref{thm:prod-pcs}). 
In the following, we provide a sketch of the algorithm for the case $\X=\Z=\Y$ and refer the readers to the detailed \cref{alg:prod}.
Intuitively, the idea is to ``break down'' the construction of the product circuit in a recursive manner by exploiting compatibility.
The base case is where $\p$ and $\q$ are input units with simple parametric forms. 
Their product can be represented as a single input unit if we can find a simple parametric form for it, which is the case, \eg for products of 
exponential families such as (multivariate) Gaussians.
Next, we consider the inductive steps where $\p$ and $\q$ are two sum or product units.

On the one hand, if $\p$ and $\q$ are compatible product units, they decompose $\X$ the same way for some ordering of inputs; \ie $\p(\X)\!=\!\p_1(\X_1)\p_2(\X_{2})$ and $\q(\X)\!=\!\q_1(\X_1)\q_2(\X_{2})$.
Then, their product $m$ as a decomposable circuit can be constructed recursively from the products of their inputs:
$m(\X) = (\p_1\q_1)(\X_1)\cdot(\p_2\q_2)(\X_2)$.

On the other hand, if $\p$ and $\q$ are smooth sum units, written as $\p(\X)\!=\!\sum_i \theta_i \p_i(\X)$ and $\q(\X)\!=\!\sum_j \theta'_j \q_j(\X)$, we can obtain their product $m$ recursively by distributing sums over products.
In other words, $m(\X)\!=\!\sum_{i,j}\theta_i\theta'_j(\p_i\q_j)(\X)$.
Note that if both input circuits are also deterministic, $m$ is
also deterministic since $\supp(\p_i \q_j)\!=\!\supp(\p_i)\cap\supp(\q_j)$ are disjoint for different $i,j$. 
Combining these, the algorithm will recursively compute the product of each pair of units in $\p$ and $\q$ with matching scopes. Assuming efficient products for input units, the overall complexity is 
$\bigO(\abs{\p}\abs{\q})$.

\subsection{Beyond Sums and Products}
\label{sec:funcs}

It is now natural to ask which other operations we can tractably apply over circuits beyond sum and products.
To formalize it, we are looking for a functional $f$ such that, given a circuit $\p(\X)$ with certain structural properties, $f(\p(\X))$ can be compactly represented as a smooth and decomposable circuit in order to admit tractable integration.

To that end, let us extract the ``ingredients for tractability'' 
from the previous section.
As usual, we can assume to apply $f$ to the input units of $\p$ and obtain tractable representations for the new input units; this is generally the case for simple parametric input functions.
Next, tractability of a function over circuits is the result of two key characteristics: that it decomposes over products and over sums.
In other words, the first condition 
is that $f(\p_1(\X_1)\cdot\p_2(\X_2))$ can be broken down to either a product $f(\p_1(\X_1))\cdot f(\p_2(\X_2))$ or sum $f(\p_1(\X_1)) + f(\p_2(\X_2))$.
Second, we want $f$ to similarly decompose over sum units; that is, $f(\p_1(\X_1)+\p_2(\X_2))$ also yields a product or sum of $f(\p_1(\X_1))$ and $f(\p_2(\X_2))$.
\begin{lemma}
\label{lem:f-properties}
Let $f$ be a continuous function over reals.
If $f(x)$ satisfies either of the above two conditions, then it must either be a linear function or take one of the following forms: $x^\beta$, $\beta\log(x)$, or $\exp(\beta\cdot x)$ for $\beta\in\R$.

\end{lemma}

As a consequence, in the following we investigate the powers, logarithms, and exponentials of circuits and complete our atlas of simple transformations.

\subsection{Powers of a Circuit}
\label{sec:pow-pc}

The $\alpha$-power of a PC $\p(\X)$ for an $\alpha\in\R$ is denoted as $\p^{\alpha}(\X)$
and is an operation needed to compute generalizations of the entropy of a PC
and related divergences (\cref{sec:queries}). 
Let us first consider natural powers ($\alpha \in \N$).
If $\p$ is only smooth and decomposable, computing the power circuit $\p^\alpha$ is \#P-hard (\cref{thm:hardness-natural-pow}). By additionally enforcing structured-decomposability, $\p^\alpha$ can be constructed by directly applying the product operation repeatedly,
which leads to the time complexity $\bigO(\abs{\p}^\alpha)$.
However, we prove in \cref{thm:hardness-natural-pow-sd} that the exponential dependence on $\alpha$ is unavoidable unless P=NP, rendering the operation intractable for large $\alpha$.

We now turn our attention to powers for a non-natural $\alpha\in\R$. As zero raised to the negative power is undefined, we instead consider the \textit{restricted $\alpha$-power}:
\begin{equation*}
    \proj{\p^{\alpha}(\x)}{\supp(\p)} = \begin{cases}
(\p(\x))^\alpha &\text{if $\x\in\supp(\p)$}\\
0 &\text{otherwise.}
\end{cases}
\end{equation*}
Note that this is equivalent to the $\alpha$-power if $\alpha\geq0$.
Abusing notation, we will also denote this by $\p^\alpha(\x)\id{\x\in\supp(\p)}$,
where $\id{\cdot}$ stands for indicator functions.
Interestingly, the power circuit in general is hard to compute even for structured-decomposable PCs.
For instance, we show in \cref{thm:inv} that building a decomposable circuit that computes the $\alpha$-power of $\p$ for $\alpha\!=\!-1$, i.e.\ its reciprocal circuit,
is \#P-hard even if $\p$ is structured-decomposable.

The key property that enables efficient computation of power circuits is determinism. More interestingly, we do not require structured-decomposability, but only smoothness and decomposability (\cref{thm:real-pow-det}).
As before, the algorithm proceeds in a recursive manner, for which a sketch is given here and details are left for \cref{appx:power}.

If $\p$ is a decomposable product unit, then its $\alpha$-power decomposes into a product of powers of its inputs:
\begin{align*}
    &\left( \p_1(\x_1)\cdot\p_2(\x_2) \right)^\alpha \id{\x\in\supp(\p)} = \p_1^\alpha(\x_1) \id{\x_1\in\supp(\p_1)} \cdot \p_2^\alpha(\x_2) \id{\x_2\in\supp(\p_2)}.
\end{align*}
The key observation above is that the support of a decomposable product unit $\p$ is simply the Cartesian product of the supports of its inputs: $\supp(\p)=\supp(\p_1)\times\supp(\p_2)$.

Next, if $\p$ is a smooth and deterministic sum unit, we can ``break down'' the computation of power over the disjoint supports carried respectively by the inputs of $\p$:
\begin{align*}
    &\left( \sum_i \theta_i \p_i(\x) \id{\x\in\supp(\p_i)}\right)^\alpha \id{\x\in\supp(\p)} =\sum_i \theta_i^\alpha \p_i^\alpha(\x)\id{\x\in\supp(\p_i)}.
\end{align*}
Here, we use the fact that for any $\x$, at most one indicator $\id{\x\in\supp(\p_i)}$ evaluates to 1.
As such,
when multiplying a deterministic sum unit with itself, each input will only have overlapping support with itself, thus effectively matching product units only with themselves.
This is why decomposability suffices.
In conclusion, this recursive decomposition of the power of a circuit will result in the power circuit having the same structure as the original circuit, with input functions and sum parameters replaced by their $\alpha$-powers.
The space and time complexity of the algorithm is $\bigO(\abs{\p})$ for smooth, deterministic, and decomposable PCs, even for natural powers. 
This will be a key insight to  compactly multiply circuits with the same support structure, such as when computing logarithms (\cref{sec:log-pc}) and entropies (\cref{sec:queries}).

We can already see an example of how simple operations are composed to derive other tractable operations. Consider the quotient of two circuits $\p(\X)$ and $\q(\X)$, defined as $\p(\X)/\q(\X)$, which often appears in queries such as KLD or Itakura-Saito divergence (\cref{sec:queries}). 
The quotient can be computed by first taking the reciprocal circuit (\ie the $(-1)$-power) of $\q$, followed by its product with $\p$.
Thus, if $\q$ is deterministic and compatible with $\p$, we can take its reciprocal---which will have the same structure as $\q$---and multiply with $\p$ to obtain the quotient as a decomposable circuit (\cref{thm:ratio-pcs}). 
We prove that the quotient between $p$ and a non-deterministic $q$ is \#P-Hard even if they are compatible (\cref{thm:hardness-ratio-pcs}).

\subsection{Logarithms of a PC}
\label{sec:log-pc}

The logarithm of a PC $\p(\X)$, denoted $\log\p(\X)$,
is fundamental for
computing 
quantities like entropies and divergences between distributions (\cref{sec:queries}).
Since the log is undefined for 
$0$ we will again consider the \textit{restricted logarithm}:
\begin{equation*}\label{eq:det-rest-log}
    \proj{\log\p(\x)}{\supp(\p)} = \begin{cases}
{\log\p(\x)} &\text{if $\x\in\supp(\p)$}\\
0 &\text{otherwise.}
\end{cases}
\end{equation*}
Unsurprisingly, computing a decomposable log circuit is a hard problem,\footnote{Note that while one usually performs computations on PCs in the log-domain for stability~\citep{peharz2019random,peharz2020einsum}, they take a logarithm of the output of a PC \textit{after} integrating some variables out;
whereas, we are interested in a compact representation of the logarithm circuit over which to perform integration.} specifically, \#P-hard even if the input circuit is smooth and structured-decomposable (\cref{thm:hardness-log}).

Again, the introduction of determinism would make the operation tractable,
by allowing it to decompose over the support of the PC, hence over its sum units.
Differently, the logarithm operation would normally turn a product unit $\p$ into a single sum unit $s$ over disjoint scopes.
To retrieve a proper smooth circuit, we join the inputs of $s$ into product units that have additional dummy inputs, \ie that outputs 1 over the corresponding missing supports and 0 elsewhere. 
For instance, consider the restricted logarithm over a product unit $\p(\X)=\p_1(\X_1)\cdot\p_2(\X_2)$, \ie $\log (\p_1(\x_1)\cdot\p_2(\x_2))\cdot\id{\x\in\supp(\p)}$.
We can represent it as the smooth sum unit
\begin{align*}
    &\left(\log (\p_{1}(\x_{1}))\cdot \id{\x_1\in\supp(\p_{1})}\right)\cdot\id{\x_2\in\supp(\p_{2})} + \left(\log (\p_{2}(\x_{2}))\cdot \id{\x_2\in\supp(\p_{2})}\right) \cdot\id{\x_1\in\supp(\p_{1})}
\end{align*}
by recalling from \cref{sec:pow-pc} 
that the support of a decomposable product unit can be written as the Cartesian product of the supports of its inputs.

\begin{figure}[!t]
    \centering
    \hspace{30pt}\includegraphics[width=.25\columnwidth,valign=t]{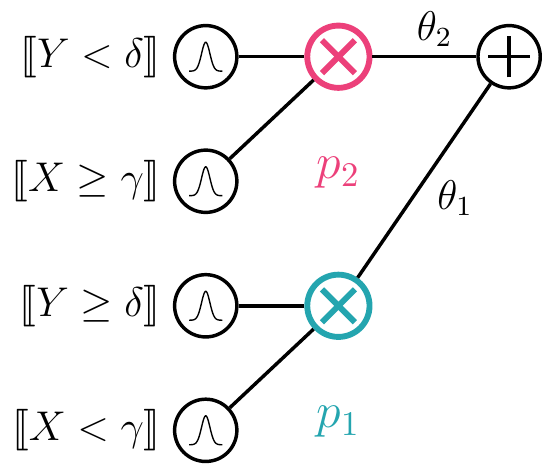}\hfill
    \includegraphics[width=.55\columnwidth,valign=t]{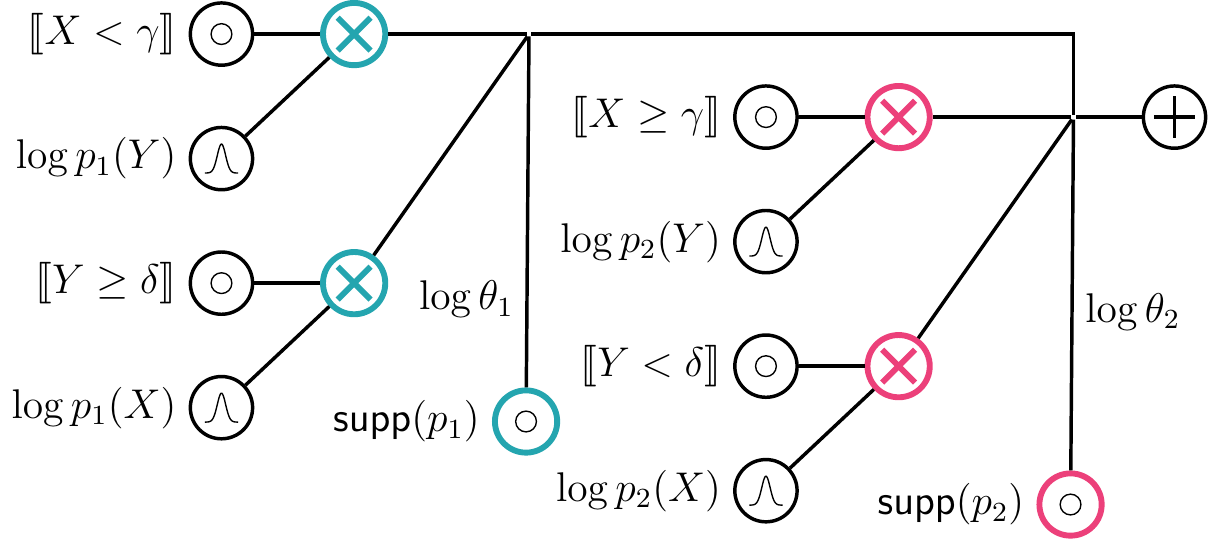}\hspace{30pt}
    \vspace{-0.4em}
    \caption{Building the logarithmic circuit (right) for a deterministic PC (left) whose input units are labeled by their supports.
    A single sum unit is introduced over smoothed product units and additional dummy input units which share the same support across circuits if they have the same color.
    }
    \label{fig:log-transform}
    \vspace{-1.0em}
\end{figure}

Breaking the logarithm over deterministic sum units follows the same idea of the restricted power and is detailed by \cref{thm:log-det}.
Ultimately, we can merge the sum units coming from the logarithm of products into a single sum 
to obtain a polysize circuit (\cref{alg:log}).
\cref{fig:log-transform} illustrates this process for a small PC.
Note that constructing a logarithm circuit in such a way would {compromise} determinism for the sum unit $s$, as multiple of its newly introduced inputs would be non-zero for a single configuration $\x$. 
Nevertheless, these inputs of $s$ can be clearly partitioned into groups sharing the same support of the original product unit $\p$, %
as illustrated in \cref{fig:log-transform}.  
This implies that whenever we have to multiply a deterministic circuit and its logarithmic circuit---for instance to compute its entropy (\cref{sec:queries})---we can leverage the sparsifying effect of non-overlapping supports and perform only a linear number of products (\cref{sec:pow-pc}).

\subsection{Exponentials of a Circuit}
\label{sec:exp-circuit}

The exponential of a circuit $\p(\X)$, \ie $\exp(\p(\X))$, is %
the inverse operation of the logarithm and is a fundamental operation when representing distributions such as log-linear models~\citep{koller2009probabilistic}. 
Similarly to the logarithm, 
building a decomposable circuit that encodes an exponential of a circuit is \#P-hard in general (\cref{thm:hardness-exp}).
Unlike the logarithm however, restricting the operation to deterministic circuits does not help with tractability, since the issue comes from product units:
the exponential of a product is 
neither a sum nor product of exponentials.

Nevertheless, if $\p$ encodes a linear sum over its variables, \ie $\p(\X)=\sum_{i}\theta_i X_i$, we could easily represent its exponential as a circuit comprising a single decomposable product unit (\cref{thm:exp-linear}).
If we were to add an additional deterministic sum unit over many 
omni-compatible circuits built in such a way,
we would retrieve a mixture of truncated exponential model~\citep{moral2001mte,zeng2020scaling}.
This is the largest class of tractable exponentials we know so far.
Enlarging its boundaries is an interesting open problem.

\section{\ldots to Complex Compositional Queries}
\label{sec:queries}

In this section, 
we show how our atlas of simple tractable operators 
can be effectively used to characterize several advanced queries, generalizing existing results in the literature and charting the tractability boundaries for  novel ones.

\definecolor{ultramarine}{RGB}{40,40,40}

\begin{table*}[!t]
    \caption{\textbf{\textit{Tractability and hardness of information-theoretic queries over circuits.}}
    Tractability given some conditions over the input circuits; computational hardness without some of these input condition. 
    } 
    \label{tab:queries}
    \vspace{-0.5em}
    \centering
    \scalebox{0.68}{
    {\renewcommand{\arraystretch}{1.0}
    \begin{tabular}{lcl@{\hspace{0.3em}}ll@{\hspace{0.3em}}l}
        \toprule
        \multicolumn{2}{c}{Query} & \multicolumn{2}{c}{Tractability Conditions} & \multicolumn{2}{c}{Hardness}
        \\ \midrule
        \textsc{Cross Entropy} & $-\textcolor{ultramarine}{\int} \p(\x) \log \q(\x)$\:d\X & Cmp, $\q$ Det & (\cref{thm:cross-entropy}) & \#P-hard w/o Det & (\cref{thm:hardness-cross-entropy}) \\
        \textsc{Shannon Entropy} & $-\textcolor{ultramarine}{\sum} \p(\x) \log\p(\x)$ & Sm, Dec, Det & (\cref{thm:entropy}) & coNP-hard w/o Det & (\cref{thm:hardness-entropy}) \\
        \multirow{2}{*}{\textsc{Rényi Entropy}} & $(1-\alpha)^{-1}\log\textcolor{ultramarine}{\int}\p^\alpha(\x)\:d\X, \alpha\in\N$ & SD & (\cref{thm:trac-reiny-natural}) & \#P-hard w/o SD & (\cref{thm:hardness-reiny-entropy-natural}) \\
              & $(1-\alpha)^{-1}\log\textcolor{ultramarine}{\int}\p^\alpha(\x)\:d\X, \alpha\in\R_{+}$ & Sm, Dec, Det & (\cref{thm:trac-reiny-real}) & \#P-hard w/o Det & (\cref{thm:hardness-reiny-entropy-real}) \\
        \textsc{Mutual Information} & $\textcolor{ultramarine}{\int} p(\x,\y)\log(p(\x,\y)/(p(\x)p(\y)))$  & Sm, SD, Det\cref{note:marg} & (\cref{thm:mi}) & coNP-Hard w/o SD & (\cref{thm:hardness-mi}) \\
        \textsc{Kullback-Leibler Div.} & $\textcolor{ultramarine}{\int} \p(\x)\log(\p(\x)/\q(\x))d\X$ & Cmp, Det & (\cref{thm:kld}) & 
        \#P-hard w/o Det & (\cref{thm:hardness-kld}) \\
        \multirow{2}{*}{\textsc{Rényi's Alpha Div.}} & $(1-\alpha)^{-1}\log\textcolor{ultramarine}{\int}\p^\alpha(\x) \q^{1-\alpha}(\x)\:d\X, \alpha\in\N$ & Cmp, $\q$ Det & (\cref{thm:trac-alpha-div}) & %
        \#P-Hard w/o Det & (\cref{thm:hard-alpha-div})\\
              & $(1-\alpha)^{-1}\log\textcolor{ultramarine}{\int}\p^\alpha(\x) \q^{1-\alpha}(\x)\:d\X, \alpha\in\R$ & Cmp, Det & (\cref{thm:trac-alpha-div}) & 
              \#P-Hard w/o Det & (\cref{thm:hard-alpha-div})\\
        \textsc{Itakura-Saito Div.} & $\textcolor{ultramarine}{\int}[\p(\x)\slash\q(\x) -\log(\p(\x)\slash\q(\x)) -1]d\:\X$ & Cmp, Det & (\cref{thm:trac-is-div}) & 
        \#P-Hard w/o Det & (\cref{thm:hardness-is-div})
        \\              
        \textsc{Cauchy-Schwarz Div.} & $-\log \frac{\textcolor{ultramarine}{\int} \p(\x)\q(\x)d\X}{\sqrt{\textcolor{ultramarine}{\int} \p^{2}(\x)d\X\textcolor{ultramarine}{\int}\q^{2}(\x)d\X}}$ & Cmp & (\cref{thm:trac-cs-div})& %
        \#P-Hard w/o Cmp & (\cref{thm:hardness-cs-div}) \\
        \textsc{Squared loss} & $\textcolor{ultramarine}{\int}{(\p(\x)-\q(\x))^{2}}d\:\X$ & Cmp & (\cref{thm:trac-sl-div}) & %
        \#P-Hard w/o Cmp & (\cref{thm:hardness-sl-div})\\
        \bottomrule
    \end{tabular}}}
\end{table*}

Given a complex query over one or more input models that involves a pipeline of operations culminating in an integration (\cref{fig:pipelines}), we can quickly devise a tractable model class for it
by inferring the sufficient conditions needed for tractably computing each operation---starting from the last one and propagating them backwards according to \cref{tab:tract-op}. 
Consider the example of computing the KLD between two distributions $p$ and $q$ mentioned in \cref{sec:intro}.
To compute integration we require a smooth and decomposable circuit (\cref{prop:trac-integration}).
Therefore, the two circuits that participate in the product, \ie $p$ and $\log(p/q)$, should be compatible (\cref{thm:prod-pcs}).
\cref{thm:log-det} tells us that the logarithm of a circuit can be tractably computed when restricted over the support of a deterministic input circuit.
Moreover, the logarithm of a structured-decomposable circuit is going to retain this property and be compatible with its input.
Therefore, we require the quotient $p/q$ to be deterministic and compatible with $p$. We know this can be obtained if both $p$ and $q$ are deterministic and compatible (\cref{thm:ratio-pcs}).
As such, we can conclude that for two deterministic and compatible PCs $p$ and $q$ we can compute their tractable KLD (\cref{thm:kld}).
In the following, we summarize analogous derivations for many different queries, as detailed in \cref{tab:queries}, for which we report also novel complexity results to complete our theoretical understanding of these operations. 
\begin{theorem}
\label{thm:info-query}
The tractability and hardness results for complex queries as reported in \cref{tab:queries} hold.
\end{theorem}

\textbf{Shannon entropy }
Recall from \cref{sec:struct-prop} that many classical tractable models are special cases of PCs with certain structural properties.
As such, all the results for general circuits will translate over these model classes.
For instance, we can tractably compute the Shannon entropy for bounded-treewidth PGMs such as Chow-Liu trees and polytrees, as they can be represented as smooth, decomposable and deterministic PCs (\cref{thm:entropy}). 
This is possible because multiplying a circuit $\p$ with its logarithm $\log\p$ can be done in linear time as the latter will share its support structure (\cref{sec:log-pc}).
Moreover, we demonstrate in \cref{thm:hardness-entropy} that computing the Shannon entropy is coNP-hard for non-deterministic PCs.
This closes an open question recently raised by \citet{shih2020probabilistic}, 
where a linear time algorithm for selective sum-product networks, a special case of deterministic and decomposable PCs, was introduced.

\textbf{R\'enyi entropy } 
For non-deterministic PCs we can employ the tractable computation of R\'enyi entropy of order $\alpha$~\citep{renyi1961measures}, which recovers Shannon Entropy for $\alpha\rightarrow 1$.
As the logarithm is taken after integration of the power circuit,
the tractability and hardness
follow directly from those of the power operation.

\textbf{Cross entropy }
As hinted by the presence of logarithm, the cross entropy is \#P-hard to compute without determinism, even for compatible PCs (\cref{thm:hardness-cross-entropy}). %
Nevertheless, we can derive the conditions for tractability using our vocabulary of simple operations. 
As it shares some sub-operations with the KLD (\cref{fig:pipelines}), the cross entropy can be tractably computed in $\bigO(\abs{\p}\abs{\q})$ if $\p$ and $\q$ are deterministic and compatible.

\textbf{Mutual information } Building on these insights, we characterize
the tractability of mutual information (MI) between sets of variables $\X$ and $\Y$ \wrt their joint distribution $\p(\X,\Y)$ encoded as a PC. 
Let the marginals $\p(\X)$ and $\p(\Y)$ be represented as PCs as well, which can be done in linear time for smooth and decomposable PCs~\citep{choi2020pc}.
Then the MI over these three PCs can be computed via a pipeline involving product, quotient, and log operators.
From \cref{tab:tract-op}, we can infer that the MI is tractable if all circuits are compatible and deterministic.\footnote{\label{note:marg}This structural property is also known as marginal determinism~\citep{choi2020pc,choi2017optimal}.
}
For non-deterministic PCs
we prove it to be coNP-Hard (\cref{thm:hardness-mi}).

\textbf{Divergences }
\citet{LiangXAI17} proposed an efficient algorithm to compute the KLD tailored for PSDDs.
This has been the only tractable divergence available for PCs so far.
We greatly extend this panorama 
by listing other tractable divergences employing the simple operations studied so far, and additionally proving their hardness for missing structural properties over their inputs.

Rényi's $\alpha$-divergences\footnote{Several alternative formulations of $\alpha$-divergences can be found in the literature such as Amari's~\citep{minka2001expectation} and Tsallis's~\citep{opper2005expectation} divergences. However, as they share the same core operations---real powers and products of circuits---our results easily extend to them as well.}~\citep{renyi1961measures} generalize several divergences such as the KLD when $\alpha\!\rightarrow\! 1$,
 Hellinger's squared divergence when $\alpha\!=\!2^{-1}$, and the $\mathcal{X}^{2}$-divergence when $\alpha\!=\!2$~\citep{gibbs2002choosing}.
They are tractable for compatible and deterministic PCs, as is the Itakura-Saito divergence which has applications in learning and signal processing~\citep{wei2001comparison}.

For non-deterministic PCs, we list the squared loss and the Cauchy-Schwarz divergence~\citep{jenssen2006cauchy}.
The latter  has applications in mixture models for approximate inference~\citep{tran2021cauchy} 
and has been derived in closed-form for mixtures of simple parametric forms like Gaussians~\citep{kampa2011closed}, Weibull and Rayligh distributions~\citep{nielsen2012closed}. 
Our results generalize them to deep mixture models~\citep{poon2011sum}.

\textbf{Expectation queries}
Among other complex queries that can be abstracted into the general form of an expectation of a circuit $f$ \wrt a PC $\p$, \ie
$\mathbb{E}_{\x\sim\p(\X)}\left[f(\x)\right]$,
there are the moments of distributions, such as means and variances.
They can be efficiently computed for any smooth and decomposable PC, as $f$ is an omni-compatible circuit (\cref{thm:mom}).
This result generalizes the moment computation for simple models such as GMMs and HMMs as they can be encoded as smooth and decomposable PCs (\cref{sec:struct-prop}).

If $f$ is the indicator function of a logical formula, the expectation 
computes its probability \wrt the distribution $\p$.
\citet{choi2015tractable} proposed an algorithm tailored to formulas $f$ over binary variables, encoded as SDDs~\citep{darwiche2011sdd} \wrt distributions that are PSDDs. 
We generalize this result to mixed continuous-discrete distributions encoded as structured-decomposable PCs that are not necessarily deterministic and to logical formulas 
in the language of satisfiability modulo theories~\citep{barrett2018satisfiability} over linear arithmetics with univariate literals (\cref{prop:log-form}). 
Lastly, if $f$ encodes  constraints over the output distribution of a deep network we retrieve the \emph{semantic loss}~\citep{xu2018semantic}.

If $f$ encodes a classifier or a regressor, then $\mathbb{E}_\p[f]$ refers to computing its expected predictions \wrt $p$~\citep{khosravi2019what}.
Our results generalize computing the expectations of decision trees and their ensembles as proposed by~\citet{khosravi2020handling} (cf.~\cref{prop:exp-pred-dtrees}) as well as those of \textit{deep regression circuits}~\citep{khosravi2019tractable}.\footnote{Despite the name, regression circuits do not conform to our definition of circuits in \cref{def:circuits}. Nevertheless, we can translate them to our format in polytime (\cref{alg:rgc-to-circuit}).}

\section{Discussion and conclusions}

In this work we introduced a unified framework to reason about tractable model classes \wrt many queries
common in probabilistic ML and AI.
Tractability is studied by rewriting complex queries as combinations of simpler operations and pushing sufficient conditions through the latter, leading to a rich atlas that can guide and inspire future research.

The most closely related work resides in the literature of
\textit{logical circuits}, which encode Boolean functions as computational graphs with AND and OR gates.
Structural properties
analogous to those we introduced for circuits (\cref{sec:struct-prop}) can be defined for logical circuits~\citep{darwiche2002knowledge} and
deterministic logical circuits can be directly translated as circuits with sums and products instead of OR and AND gates. %
Tractable logical circuit operations such as disjunctions and conjunctions---the analogous to our (deterministic) sum and products---have been investigated for several logical formalisms~\citep{darwiche2002knowledge}.
Our results generalize the Boolean case for these operations. We also introduce novel operations, including powers and logarithms as well as complex queries such as divergences, that have no direct counterpart in the logical domain. 

Algorithms to tractably multiply two probabilistic models have been proposed in the context of probabilistic decision graphs (PDGs)~\citep{jaeger2004probabilistic,jaeger2006learning} first and PSDDs later~\citep{shen2016tractable}.
Despite the different syntax, both PDGs and PSDDs can be represented as structured-decomposable and deterministic circuits in our language~\citep{choi2020pc}.
Differently from our treatment in \cref{sec:struct-prop}, PDGs and PSDDs define compatibility in terms of special notions of hierarchical scope partitioning, namely pseudo forests~\citep{jaeger2004probabilistic} and vtrees~\citep{pipatsrisawat2008new}, respectively.
In particular, they differ from our general characterization in
that they (1) enforce a positional ordering over the partitions and (2) imbue this ordering with the semantic of conditioning over one set of variables to obtain the distribution over the others.
As such, these representations \textit{entangle determinism and compatibility}.
As we showed in \cref{thm:prod-pcs}, compatibility is sufficient for tractable multiplication, and as discussed in the previous section many algorithms tailored for PSDDs~\citep{choi2015tractable,shen2016tractable,khosravi2019tractable} can be generalized to \emph{non-deterministic} distributions in our framework.

Our property-driven analysis closes many open questions about the tractability and hardness of queries for many model classes that are special cases of circuits.
Nevertheless, other interesting questions remain open and constitute possible future directions.
For instance, demonstrating unconditional lower bounds for our representations
or extending our analysis to queries involving maximization---that is, MAP inference over probability distributions.
On the other hand, our atlas could support the design of learning routines for circuits in different ways.
First, existing algorithms~\citep{rahman2014cutset,vergari2015simplifying,peharz2019random,dang2020strudel} could be enriched by our new transformations to generate tractable structures.
Second, our analysis could help design novel algorithms to learn circuits that are tailored to answer multiple queries efficiently at once, in a sort of multi-objective optimization scenario where the algorithm trades-off circuit sizes across different queries.

\scalebox{0.01}{Io so. Ma non ho le prove.}

\section*{Acknowledgments}
The authors would like to thank Yujia Shen and Arthur Choi for insightful discussions about the product algorithm for PSDDs and Zhe Zeng for proofreading an initial version of this work.
This work is partially supported by NSF grants \#IIS-1943641, \#IIS-1633857, \#CCF-1837129, DARPA grant \#N66001-17-2-4032, a Sloan Fellowship, Intel, and Facebook.
The research of ST was partially supported by TAILOR, a project funded by EU Horizon 2020 research and innovation programme under GA No 952215.

\bibliography{refs}
\bibliographystyle{icml2021}

\clearpage
\appendix

\allowdisplaybreaks

\section{Useful Sub-Routines}

This section introduces the algorithmic construction of gadget circuits that will be adopted in our proofs of tractability as well as hardness. 
We start by introducing three primitive functions for constructing circuits---\inputUnit, \sumUnit, and \prodUnit.

$\bullet$ $\inputUnit(l_\p, \phi(\p))$ constructs an input unit that encodes a parameterized function $l_\p$ over variables $\phi(\p)$. For example, $\inputUnit(\id{X=\mathrm{True}}, X)$ and $\inputUnit(\id{X=\mathrm{False}}, X)$ represent the positive and negative literals of a Boolean variable $X$, respectively.
On the other hand, $\inputUnit(\mathcal{N}(\mu, \sigma), X)$ defines a Gaussian pdf with mean $\mu$ and standard deviation $\sigma$ as an input function.

$\bullet$ $\sumUnit(\{\p_i\}_{i=1}^{k}, \{\theta_i\}_{i=1}^{k})$ constructs a sum unit that represents the weighted combination of circuit units $\{\p_i\}_{i=1}^{k}$ encoded as an ordered set w.r.t. the correspondingly ordered weights $\{\theta_i\}_{i=1}^{k}$.

$\bullet$ $\prodUnit(\{\p_i\}_{i=1}^{k})$ builds a product unit that encodes the product of circuit units $\{\p_i\}_{i=1}^{k}$.

\subsection{Support circuit of a deterministic circuit}
\label{appx:algo-support}

\begin{algorithm}[tb]
   \caption{\suppCirc($\p, \mathsf{cache}$) 
   }
   \label{alg:support-det}
   \begin{algorithmic}[1]
   \STATE {\bfseries Input:} a smooth, deterministic, and decomposable circuit $\p$ over variables $\X$ and a cache for memorization 
   \STATE {\bfseries Output:} a smooth, deterministic, and decomposable circuit $s$ over $\X$ encoding $s(\x)=\id{\x\in\supp(\p)}$
   \LineIf{$\p \in \mathsf{cache}$}{\textbf{return} $\mathsf{cache}(\p)$}
   \IF{$\p$ is an input unit}
   \STATE $s\leftarrow\inputUnit(\id{\x\in\supp(\p)}, \scope(p))$
   \ELSIF{$\p$ is a sum unit}
   \STATE $s\leftarrow \sumUnit(\{\suppCirc(\p_i, \mathsf{cache})\}_{i=1}^{|\ch(\p)|}, \{1\}_{i=1}^{|\ch(\p)|})$
   \ELSIF{$\p$ is a product unit}
   \STATE $s\leftarrow \prodUnit(\{\suppCirc(\p_i, \mathsf{cache})\}_{i=1}^{|\ch(\p)|}|)$
   \ENDIF
   \STATE $\mathsf{cache}(\p)\leftarrow s$
   \STATE \textbf{return} $s$
\end{algorithmic}
\end{algorithm}

Given a smooth, decomposable, and deterministic circuit $\p(\X)$, its support circuit $s(\X)$ is a smooth, decomposable, and deterministic circuit that evaluates $1$ iff the input $\x$ is in the support of $\p$ (i.e., $\x \in \supp(\p)$) and otherwise evaluates $0$, as defined below.

\begin{definition}[Support circuit]
Let $\p$ be a smooth, decomposable, and deterministic PC over variables $\X$.  Its support circuit is the circuit $s$ that computes $s(\x)=\id{\x\in\supp(p)}$, obtained by replacing every sum parameter of $\p$ by 1 and every input distribution $l$ by the function $\id{\x\in\supp(l)}$. 
\end{definition}

A construction algorithm for the support circuit is provided in \cref{alg:support-det}. This algorithm will later be useful in defining some circuit operations such as the logarithm.

\subsection{Circuits encoding uniform distributions}
\label{appx:unif-pc}

We can build a 
deterministic and omni-compatible PC that encodes a (possibly unnormalized) uniform distribution over binary variables $\X=\{X_{1}, \ldots, X_{n}\}$: \ie $p(\x)=c$ for a constant $c\in\R_{+}$ for all $\x\in\val(\X)$. 
Specifically, $\p$ can be defined as a single sum unit with weight $c$ that receives input from a product unit over $n$ univariate input distribution units that always output 1 for all values $\val(X_i)$. This construction is summarized in \cref{alg:uniform-pc}. It is a key component in the algorithms for many tractable circuit transformations/queries as well as in several hardness proofs.

\begin{algorithm}[!tb]
   \caption{\textsc{uniformCircuit}($\X, c$) 
   }
   \label{alg:uniform-pc}
   \begin{algorithmic}[1]
   \STATE {\bfseries Input:} a set of variables $\X$ and constant $c\in\R_{+}$. 
   \STATE {\bfseries Output:} a deterministic and omni-compatible PC encoding an unnormalized uniform distribution over $\X$.
   \STATE $n \leftarrow \{\}$
   \FOR{$i=1$ \textbf{to} $|\X|$}
   \STATE $m \leftarrow \{\}$
   \FOR{$x_i$ \textbf{in} $\val (X_i)$}
   \STATE $m \leftarrow m \cup \{\inputUnit(\id{X_i=x_i},X_i)\}$ 
   \ENDFOR
   \STATE $n \leftarrow n \cup \{\sumUnit(m, \{1\}_{j=1}^{|\val(X_i)|})\}$
   \ENDFOR
   \STATE \textbf{return} $\sumUnit( \{\prodUnit(n)\}, \{c\})$
\end{algorithmic}
\end{algorithm}

\subsection{A circuit representation of the \textsf{\#3SAT} problem}
\label{appx:3SAT-circuit}

We define a circuit representation of the \textsf{\#3SAT} problem, following the construction in \citet{khosravi2019tractable}. Specifically, we represent each instance in the \textsf{\#3SAT} problem as two poly-sized structured-decomposable and deterministic circuits $\p_\beta$ and $\p_\gamma$, such that the partition function of their product equals the solution of the original \textsf{\#3SAT} problem.

\textsf{\#3SAT} is defined as follows: given a set of $n$ boolean variables $\X = \{X_1, \dots, X_n\}$ and a CNF that contains $m$ clauses $\{c_1, \dots, c_m\}$ (each clause contains exactly 3 literals), count the number of satisfiable worlds in $\val(\X)$.

For every variable $X_i$ in clause $c_j$, we introduce an auxiliary variable $X_{ij}$. Intuitively, $\{X_{ij}\}_{j=1}^{m}$ are copies of the variable $X_i$, one for each clause. Therefore, for any $i$, $\{X_{ij}\}_{j=1}^{m}$ share the same value (i.e., true or false), which can be represented by the following formula $\beta$:
\begin{align*}
    \beta \equiv \bigwedge\limits_{i=1}^{n} (X_{i1} \Leftrightarrow X_{i2} \Leftrightarrow \cdots \Leftrightarrow X_{im}).
\end{align*}

Then we can encode the original CNF in the following formula $\gamma$ by substituting $X_i$ with the respective $X_{ij}$ in each clause:
\begin{align*}
    \gamma \equiv \bigwedge\limits_{j=1}^{m} \bigvee\limits_{i: X_i \in \phi(c_j)} l(X_{ij}),
\end{align*}
\noindent where $\phi (c)$ denotes the variable scope of clause $c$, and $l(X_{ij})$ denotes the literal of $X_i$ in clause $c_j$. Since $\beta$ restricts the variables $\{X_{ij}\}_{j=1}^{m}$ to have the same value, the model count of $\beta \wedge \gamma$ is equal to the model count of the original CNF. 

We are left to show that both $\beta$ and $\gamma$ can be compiled into a poly-sized 
structured-decomposable and deterministic circuit. We start from compiling $\beta$ into a circuit $\p_\beta$. 
Note that for each $i$, $(X_{i1} \Leftrightarrow \cdots \Leftrightarrow X_{im})$ has exactly two satisfiable variable assigments (i.e., all true or all false), it can be compiled as a sum unit $a_i$ over two product units $b_{i1}$ and $b_{i2}$ (both weights of $a$ are set to $1$), where $b_{i1}$ takes inputs from the positive literals $\{X_{i1}, \dots, X_{im}\}$ and $b_{i2}$ from the negative literals $\{\neg X_{i1}, \dots, \neg X_{im}\}$. 
Then $\p_\beta$ is represented by a product unit over $\{a_1, \dots, a_n\}$. Note that by definition this $\p_\beta$ circuit is structured-decomposable and deterministic.

We proceed to compile $\gamma$ into a polysized structured-decomposable and deterministic circuit $\p_\gamma$. 
Note that in \textsf{\#3SAT}, each clause $c_j$ contains 3 literals. Therefore, for any $j \in \{1, \dots, m\}$, $\bigvee_{X_i \in \phi (c_j)} l(X_{ij})$ has exactly 7 models w.r.t. the variable scope $\phi (c_j)$. 
Hence, we compile $\bigvee_{X_i \in \phi (c_j)} l(X_{ij})$ into a circuit $d_j$, which is a sum unit with 7 inputs $\{e_{j1}, \dots, e_{j7}\}$. Each $e_{jh}$ 
is constructed as a product unit over variables $\{X_{1j}, \dots, X_{nj}\}$ that represents the $h$-th model of clause $c_j$. 
More formally, we have
    $e_{jh} \leftarrow \prodUnit(\{g_{ijh}\}_{i=1}^{n})$,
\noindent where $g_{ijh}$ is a sum unit over literals $X_{ij}$ and $\neg X_{ij}$ (with both weights being $1$) if $i \not\in \phi(c_j)$ and otherwise $g_{ijh}$ is the literal unit corresponds to the $h$-th model of clause $c_j$. The circuit $\p_\gamma$ representing the formula $\gamma$ is constructed by a product unit with inputs $\{d_j\}_{j=1}^{m}$. By construction this circuit is also structured-decomposable and deterministic.

\section{Circuit Operations}

This section formally presents the tractability and hardness results w.r.t. circuit operations summarized in \cref{tab:tract-op}---sums, products, quotients, powers, logarithms, and exponentials. 
For each circuit operation, we provide both its proof of tractability by constructing a polytime algorithm given sufficient structural constraints and novel hardness results that identify necessary structural constraints for the operation to yield a decomposable circuit as output.

Throughout this paper, we will show hardness of operations to output a decomposable circuit by proving hardness of computing the partition function of the output of the operation.
This follows from the fact that we can smooth and integrate a decomposable circuit in polytime, thereby making the former problem at least as hard as the latter.

For the tractability theorems, we will assume that the operation referenced by the theorem is tractable over input units of circuit or pairs of compatible input units whose element belong each to a circuit.
For example, for \cref{thm:prod-pcs} we assume tractable product of input units sharing the same scope and for \cref{thm:real-pow-det} we assume that the powers of the input units can be tractably represented as a single new unit.

Moreover, in the following results, we will adopt a more general definition of compatibility that can be applied to circuits with different variable scopes, which is often useful in practice.
Formally, consider two circuits $\p$ and $\q$ with variable scope $\Z$ and $\Y$. Analogous to \cref{def:compatibility}, we say that $\p$ and $\q$ are compatible over variables $\X = \Z \cap \Y$ if (1) they are smooth and decomposable and (2) any pair of product units $n\in\p$ and $m\in\q$ with the same overlapping scope with $\X$ can be rearranged into mutually compatible binary products.
Note that since our tractability results hold for this extended definition of compatibility, they are also satisfied under \cref{def:compatibility}.

\subsection{Sum of Circuits}

The hardness of the sum of two circuits to yield a deterministic circuit has been proven by \citet{shen2016tractable} in the context of arithmetic circuits (ACs)~\citep{darwiche2002knowledge}.
ACs can be readily turned into circuits over binary variables according to our definition by translating their input parameters into sum parameters as done in~\citet{rooshenas2014learning}.

A sum of circuits will preserve decomposability and related properties as the next proposition details.
\begin{proposition}[Closure of sum of circuits]
\label{thm:trac-sums}
Let $\p(\Z)$ and $\q(\Y)$ be decomposable circuits. Then their sum circuit $s(\Z\cup\Y)=\theta_1\cdot\p(\Z)+\theta_2\cdot\q(\Y)$ for two reals $\theta_1,\theta_2\in\R$ is decomposable. If $\p$ and $\q$ are structured-decomposable and compatible, then $s$ is structured-decomposable and compatible with both $\p$ and $\q$. Lastly, if both inputs are also smooth, $s$ can be smoothed in polytime. 
\begin{proof}
If $\p$ and $\q$ are decomposable, $s$ is also decomposable by definition (no new product unit is introduced). 
If they are also structured-decomposable and compatible, $s$ would be structured-decomposable and compatible with $\p$ and $\q$ as well, as summation does not affect their hierarchical scope partitioning.
Note that if one input is decomposable and the other omni-compatible, then $s$ would only be decomposable.

If $\Z=\Y$ then $s$ would be smooth; otherwise we can smooth it in polytime~\citep{darwiche2009modeling,shih2019smoothing}, \ie by realizing the circuit
\begin{align*}
    \label{eq:sum-smooth}
    s(\x) =& \;\theta_{1}\cdot\p(\z)\cdot\id{\q(\proj{\x}{\Y\setminus\Z})\neq 0} +\\ &\;\theta_{2}\cdot\q(\y)\cdot\id{\p(\proj{\x}{\Z\setminus\Y})\neq 0}
\end{align*}
where $\id{\q(\proj{\x}{\Y\setminus\Z})\neq 0}$ (resp.~$\id{\p(\proj{\x}{\Z\setminus\Y})\neq 0}$ ) can be encoded as an input distribution over variables $\Y\setminus\Z$ (resp.$\Z\setminus\Y$).
Note that if the supports of $\p(\Z\setminus\Y)$ and $\q(\Y\setminus\Z)$ are not bounded, then integrals over them would be unbounded as well.
\end{proof}
\end{proposition}

\subsection{Product of Circuits}

\begin{theorem}[Hardness of product of circuits]
\label{thm:prod-dec}
Let $\p$ and $\q$ be two structured-decomposable and deterministic circuits over variables $\X$. Computing  their product $m(\X)=\p(\X)\cdot\q(\X)$ as a decomposable circuit is \#P-Hard.\footnote{Note that this implies that product of decomposable circuits is also \#P-hard, as decomposability is a weaker condition than structured-decomposability. The hardness results throughout this paper translate directly when input properties are relaxed.}
\end{theorem}

\begin{proof}

As noted earlier, we will prove hardness of computing the product by showing hardness of computing the partition function of a product of two circuits.
In particular, let $\p$ and $\q$ be two structured-decomposable and deterministic circuits over binary variables $\X$. Then, computing the following quantity is \#P-hard:
\begin{align*}
    \sum_{\x \in \val(\X)} \p(\x) \cdot \q(\x) \tag{\textsf{MULPC}}.
\end{align*}

The following proof is adapted from the proof of Thm.\ 2 in \citet{khosravi2019tractable}.
We reduce the \textsf{\#3SAT} problem defined in \cref{appx:3SAT-circuit}, which is known to be \#P-hard, to \textsf{MULPC}.
Recall that $\p_\beta$ and $\p_\gamma$, as constructed in \cref{appx:3SAT-circuit}, are structured-decomposable and deterministic; additionally, the partition function of $\p_\beta \cdot \p_\gamma$ is the solution of the corresponding \textsf{\#3SAT} problem.
In other words, computing \textsf{MULPC} of two structured-decomposable and deterministic circuits $\p_\beta$ and $\p_\gamma$ exactly solves the original \textsf{\#3SAT} problem.
Therefore, computing the product of two structured-decomposable and deterministic circuits is \#P-Hard.
\end{proof}

\begin{theorem}[Tractable product of circuits]
\label{thm:prod-pcs}
Let $\p(\Z)$ and $\q(\Y)$ be two compatible circuits over variables $\X=\Z\cap\Y$.
Then,  computing their product $\m(\X)=\p(\Z)\cdot\q(\Y)$ as a decomposable circuit can be done in $\bigO(\abs{\p}\abs{\q})$ time and space.
If both $\p$ and $\q$ are also deterministic, then so is $\m$,
moreover if $\p$ and $\q$ are structured-decomposable then $\m$ is compatible with $\p$ (and $\q$) over $\X$.
\end{theorem}

\begin{proof}

The proof proceeds by showing that computing the product of (i) two smooth and compatible sum units $\p$ and $\q$ 
and (ii) two smooth and compatible product units $\p$ and $\q$ 
given the product circuits w.r.t. pairs of child units from $\p$ and $\q$ (i.e., $\forall r\in\ch(\p)\, s\in\ch(\q), (r \!\cdot\! s)(\X)$) takes time $\bigO(\abs{\ch(\p)} \abs{\ch(\q)})$. Then, by recursion, the overall complexity in time and space are both $\bigO(\abs{\p}\abs{\q})$.
\cref{alg:prod} illustrates the overall process in detail.

If $\p$ and $\q$ are two sum units defined as $\p(\x)=\sum_{i\in\ch(\p)}\theta_{i}\p_{i}(\x)$ and $\q(\x)=\sum_{j\in\ch(\q)}\theta^\prime_{j}\q_{j}(\x)$, respectively. Then, their product $m(\x)$ can be broken down to the weighted sum of $\abs{\ch(\p)}\!\cdot\!\abs{\ch(\q)}$ circuits that represent the products of pairs of their inputs:
\begin{align*}
    m(\x) & = \left ( \sum_{i\in\ch(\p)}\theta_{i}\p_{i}(\x) \right ) \left ( \sum_{j\in\ch(\q)}\theta^\prime_{j}\q_{j}(\x) \right ) \\
    = & \sum_{i\in\ch(\p)} \sum_{j\in\ch(\q)} \theta_{i} \theta^\prime_{j} (\p_i \q_j)(\x).
\end{align*}
Note that this Cartesian product of units is a deterministic sum unit if both $\p$ and $\q$ were deterministic sum units, as $\supp(\p_i \q_j)\!=\!\supp(\p_i)\cap\supp(\q_j)$ are disjoint for different $i,j$. 

If $\p$ and $\q$ are two product units defined as $\p(\X)=\p_1(\X_1)\p_2(\X_2)$ and $\q(\X)=\q_1(\X_1)\q_2(\X_2)$, respectively.
Then, their product $m(\x)$ can be constructed recursively from the product of their inputs:
\begin{align*}
    m(\x) & = \p_1(\x_1)\p_2(\x_2) \cdot \q_1(\x_1) \q_2(\x_2) \\
    & = \p_1(\x_1)\q_1(\x_1) \cdot \p_2(\x_2) \q_2(\x_2) \\
    & = (\p_1 \q_1) (\x_1) \cdot (\p_2 \q_2) (\x_2).
\end{align*}
Note that by this construction $\m$ retains the same scope partitioning of $\p$ and $\q$, hence if they were structured-decomposable, $\m$ will be structured-decomposable and compatible with $\p$ and $\q$.  
\end{proof}

\begin{algorithm}[!tb]
   \caption{\textsc{multiply}($\p, \q, \mathsf{cache}$)}
   \label{alg:prod}
   \begin{algorithmic}[1]
   \STATE {\bfseries Input:} two 
   circuits $\p(\Z)$ and $\q(\Y)$ that are compatible over $\X=\Z\cap\Y$ and a cache for memoization
   \STATE {\bfseries Output:} their product circuit $\m(\Z\cup\Y)=\p(\Z)\q(\Y)$
   \LineIf{$(\p, \q) \in \mathsf{cache}$}{\textbf{return} $\mathsf{cache}(\p, \q)$}
   \IF{$\scope(\p)\cap\scope(\q)=\emptyset$}
        \STATE  $\m\leftarrow\prodUnit(\{\p, \q\}); \; s \leftarrow \mathrm{True}$ 
   \ELSIF{$\p,\q$ are input units}
        \STATE  $\m\leftarrow\inputUnit(\p(\Z)\cdot \q(\Y), \Z\cup\Y)$
        \STATE $s\leftarrow\id{\supp(\p(\X))\cap\supp(\q(\X))\neq\emptyset}$ 
    \ELSIF{$\p$ is an input unit}
       \STATE $\node \leftarrow \{\}; s \leftarrow \mathrm{False}$ $\slash\slash \q(\Y)=\sum_{j}\theta^\prime_j\q_j(\Y)$
       \FOR{$j=1$ \textbf{to} $|\ch(\q)|$}
        \STATE $\node', s'\leftarrow\text{\textsc{multiply}}(\p, \q_j, \mathsf{cache})$
        \STATE  $\node\leftarrow\node \cup \{\node'\};\; s\leftarrow s \lor s'$
        \ENDFOR
        \LineIfElse{$s$}{$\m\leftarrow\sumUnit(\node, \{\theta^\prime_j\}_{j=1}^{|\ch(\q)|})$}{$\m\leftarrow null$}
    \ELSIF{$\q$ is an input unit}
        \STATE $\node \leftarrow \{\}; s\leftarrow\mathrm{False}$ $\slash\slash \p(\Z)=\sum_{i}\theta_i\p_i(\Z)$
        \FOR{$i=1$ \textbf{to} $|\ch(\p)|$}
        \STATE $\node', s'\leftarrow\text{\textsc{multiply}}(\p_i, \q, \mathsf{cache})$
        \STATE  $\node\leftarrow\node \cup \{\node'\};\; s\leftarrow s \lor s'$
        \ENDFOR
        \LineIfElse{$s$}{$\m\leftarrow\sumUnit(\node, \{\theta_i\}_{i=1}^{|\ch(\p)|})$}{$\m\leftarrow null$}
    \ELSIF{$\p,\q$ are product units}
       \STATE $\node \leftarrow \{\}; s\leftarrow\mathrm{True}$
      \STATE $\{\p_i, \q_i\}_{i=1}^{k}\leftarrow\mathsf{sortPairsByScope}(\p, \q, \X)$  
       \FOR{$i=1$ \textbf{to} $k$}
       \STATE $\node', s'\leftarrow\text{\textsc{multiply}}(\p_i, \q_i, \mathsf{cache})$
       \STATE  $\node\leftarrow\node \cup \{\node'\};\; s\leftarrow s\land s'$
       \ENDFOR
       \LineIfElse{$s$}{$\m\leftarrow\prodUnit(\node)$}{$\m\leftarrow null$}
    \ELSIF{$\p,\q$ are sum units}
       \STATE $\node \leftarrow \{\};\; w\leftarrow\{\};\; s\leftarrow\mathrm{False}$
       \FOR{$i=1$ \textbf{to} $|\ch(\p)|$, $j=1$ \textbf{to} $|\ch(\q)|$}
       \STATE $\node', s'\leftarrow\text{\textsc{multiply}}(\p_i, \q_j, \mathsf{cache})$
       \STATE $\node\leftarrow\node\cup\node'; w\leftarrow w\cup\{\theta_i\theta^\prime_j\}; s\leftarrow s\lor s'$
       \ENDFOR
       \LineIfElse{$s$}{$\m\leftarrow\sumUnit(\node,w)$}{$\m\leftarrow null$}
   \ENDIF
   \STATE $\mathsf{cache}(\p, \q)\leftarrow(\m, s)$
   \STATE \textbf{return} $\m, s$
\end{algorithmic}
\end{algorithm}

Possessing additional structural constrains can lead to sparser output circuits as well as efficient algorithms to construct them.
First, if one among $\p$ and $\q$ is omni-compatible, it suffices that the other is just decomposable to obtain a tractable product, whose size this time is going to be linear in the size of the decomposable circuit.
\begin{corollary}
\label{cor:omni-prod}
Let $\p$ be a smooth and decomposable circuit over $\X$ and $\q$ an omni-compatible circuit over $\X$ comprising a sum unit with $k$ inputs, hence its size is $k\abs{\X}$. 
Then, $\m(\X)=\p(\X)\q(\X)$ is a smooth and decomposable circuit constructed in $\bigO(k\abs{\p})$ time and space.
\end{corollary}

Second, if $\p$ and $\q$ have inputs with restricted supports, their product is going to be sparse, \ie only a subset of their inputs is going to yield a circuit that does not constantly output zero.
Note that in \cref{alg:prod} we can check in polytime if the supports of two units to be multiplied are overlapping by a depth-first search (realized with a Boolean indicator $s$ in \cref{alg:prod}), thanks to decomposability.
Therefore, for two compatible sum units $p$ and $q$ we will effectively build a number of units that is
\begin{align*}
    \bigO(\abs{ \{(\p_i, \q_j)|\p_i\!\in\!\ch(\p),\q_i\!\in\!\ch(\q), \supp(\p_i)\!\cap\!\supp(\q_j)\!\neq\!\emptyset\} }).
\end{align*}

In practice, this sparsifying effect will be more prominent when both $\p$ and $\q$ are deterministic.
This is because having disjoint supports is required for deterministic circuits.
This ``decimation'' of product units will be maximum if $p$ and $q$ partition the support in the very same way, for instance when we have $p=q$, \ie we are multiplying one circuit with itself, or we are dealing with a logarithmic circuit (cf.~\cref{sec:log-pc}).
In such a case, we can omit the depth-first check for overlapping supports of the product units participating in the product of a sum unit.
If both $\p$ and $\q$ have an identifier for their supports, we can simply check for equality of their identifiers.
This property and algorithmic insight will be key when computing powers of a deterministic circuit and its entropies (cf.~\cref{appx:entropies}), as it would suffice the input circuit $p$ to be decomposable (cf.~\cref{sec:funcs}) to obtain a linear time complexity.

\begin{algorithm}[!tb]
   \caption{\textsc{sortPairsByScope}($\p, \q, \X$)}
   \label{alg:sort-pairs-by-scope}
   \begin{algorithmic}[1]
   \STATE {\bfseries Input:} two decomposable and compatible product units $p$ and $q$, and a variable scope $\X$. %
   \STATE {\bfseries Output:} Pairs of compatible sum units $\{(p_i, q_i)\}_{i = 1}^{k}$.
   \STATE $\mathrm{children}\_p \leftarrow \{p_i\}_{i=1}^{\abs{\ch(\p)}}, \quad \mathrm{children}\_q \leftarrow \{q_i\}_{i=1}^{\abs{\ch(\q)}}$
   \STATE $\mathrm{pairs} \leftarrow \{\}$. $\slash\slash$ ``$\mathrm{pairs}$'' stores circuit pairs with matched scope.
   \STATE $\mathrm{cmp}\_p \leftarrow \{\{\}\}_{i=1}^{\abs{\ch(\p)}}, \quad \mathrm{cmp}\_q \leftarrow \{\{\}\}_{j=1}^{\abs{\ch(\q)}}$.\\ $\slash\slash$ $\mathrm{cmp}\_p[i]$ (resp. $\mathrm{cmp}\_q[j]$) stores the children of $\q$ (resp. $\p$) whose scopes are subsets of $\p_i$'s (resp. $\q_j$'s) scope.
   \FOR{i = 1 \textbf{to} $|\ch(\p)|$}
   \FOR{j = 1 \textbf{to} $|\ch(\q)|$}
   \IF{$\phi(\p_i)\cap\X = \phi(\q_j) \cap\X$}
   \STATE $\mathrm{pairs}.append((\p_i, \q_j))$
   \STATE $\mathrm{children}\_p.pop(\p_i), \quad \mathrm{children}\_q.pop(\q_j)$
   \ELSIF{$\phi(\p_i)\cap\X \subset \phi(\q_j) \cap\X$}
   \STATE $\mathrm{cmp}\_q[j].append(\p_i)$
   \STATE $\mathrm{children}\_p.pop(\p_i), \quad \mathrm{children}\_q.pop(\q_j)$
   \ELSIF{$\phi(\q_j)\cap\X \subset \phi(\p_i)\cap\X$}
   \STATE $\mathrm{cmp}\_p[i].append(\q_j)$
   \STATE $\mathrm{children}\_p.pop(\p_i), \quad \mathrm{children}\_q.pop(\q_j)$
   \ENDIF
   \ENDFOR
   \ENDFOR
   
   \FOR{$i=1$ \textbf{to} $\abs{\ch(\p)}$}
   \IF{$len(\mathrm{cmp}\_p[i]) \neq 0$}
   \STATE $s \leftarrow \sumUnit( \{\prodUnit(\mathrm{cmp}\_p[i])\}, \{1\})$
   \STATE $\mathrm{pairs}.append((\p_i, s))$
   \ENDIF
   \ENDFOR
   
   \FOR{$j=1$ \textbf{to} $\abs{\ch(\q)}$}
   \IF{$len(\mathrm{cmp}\_q[j]) \neq 0$}
   \STATE $r \leftarrow \sumUnit( \{\prodUnit(\mathrm{cmp}\_q[j])\}, \{1\})$
   \STATE $\mathrm{pairs}.append((r, \q_j))$
   \ENDIF
   \ENDFOR
   
   \FOR{$r, s$ \textbf{in} $zip(\mathrm{children}\_p, \mathrm{children}\_q)$}
   \STATE $\mathrm{pairs}.append((r, s))$
   \ENDFOR
   \IF{$len(\mathrm{children}\_p) > len(\mathrm{children}\_q)$}
   \FOR{$i=len(\mathrm{children}\_q)+1$ \textbf{to} $len(\mathrm{children}\_p)$}
   \STATE $\mathrm{pairs}.append((\mathrm{children}\_p[i], \mathrm{children}\_q[1]))$
   \ENDFOR
   \ELSIF{$len(\mathrm{children}\_p) < len(\mathrm{children}\_q)$}
   \FOR{$j=len(\mathrm{children}\_p)+1$ \textbf{to} $len(\mathrm{children}\_q)$}
   \STATE $\mathrm{pairs}.append((\mathrm{children}\_p[1], \mathrm{children}\_q[j]))$
   \ENDFOR
   \ENDIF
   \STATE \textbf{return} $pairs$
\end{algorithmic}
\end{algorithm}

\subsection{Tractable functions of circuits}

We restate the Lemma to separate the possible cases.

{\itshape {\bfseries  \cref{lem:f-properties}}
Let $f$ be a continuous function. 
If  (1) $f:\R\to\R$ satisfies $f(x + y)=f(x) + f(y)$ then it is a linear function $\beta \cdot x$;
if (2) $f:\R_{+}\to\R_{+}$ satisfies $f(x \cdot y)=f(x) \cdot f(y)$, then it takes the form $x^\beta$;
if (3)  instead  $f:\R_{+}\to\R$ satisfies $f(x \cdot y)=f(x) + f(y)$, then it takes the form  $\beta\log(x)$; and if (4) $:\R\to\R_{+}$  satisfies that $f(x + y)=f(x) \cdot f(y)$ then it is of the form $\exp(\beta\cdot x)$, for a certain $\beta\in\R$.
}

\begin{proof}
The proof of all properties follows from constructing $f$ such that we obtain a \textit{Cauchy functional equation}~\citep{jurkat1965cauchy,sahoo2011introduction}.

The condition (1) exactly takes the form of a Cauchy functional equation, then it must hold that $f(x)= \beta\cdot x$. 

For condition (2), let $g(x) = \log(f(\exp(x)))$ for all $x\in\R$, which is continuous because $f$ is.
Then, it follows that
\begin{align*}
    g(x + y) &= \log(f(\exp(x+ y)))\\ &= \log(f(\exp(x)\cdot\exp(y)))\\ &= \log(f(\exp(x)))+\log(f(\exp(y)))\\ &= g(x) + g(y)
\end{align*}
Therefore, $g(x)$ assumes the Cauchy functional form and, as in case (1), it is equal to $\beta\cdot x$.  $\beta$ can be retrieved by solving $\beta \cdot x = \log(f(\exp(x)))$ for $x=1$.  This gives $\beta=\log(f(e))$. Applying the definition of $g$, we can hence write
$$f(\exp(x))=e^{g(x)}= e^{\beta\cdot x}= \left(e^{x}\right)^{\beta}$$
Let $y\in\R_+$.  Using the identity $y=e^{\log(y)}$ it follows that:
\begin{align*}
    f(y) = f(e^{\log(y)}) = \left(e^{\log(y)}\right)^{\beta} = y^{\beta}.
\end{align*}

Condition (3) follows an analogous pattern.
Let $g(x) = f(\exp(x))$ for all $x\in\R$, which is continuous as $f$ is. 
Once again, $g$ satisfies the Cauchy functional form:
\begin{align*}
    g(x + y) &= f(\exp(x+ y)) = f(\exp(x)\cdot\exp(y))\\ &= f(\exp(x))+f(\exp(y))= g(x) + g(y)
\end{align*}
Therefore, $g(x)$ must be of the form $\beta\cdot x$ for $\beta=f(e)$.
Hence, 
$f(y) = \beta\log(y)$. 

Lastly, for condition (4), $g(x) = \log(f(x))$ for all $x\in\R$, which is continuous if $f$ is.
Then, we can retrieve the Cauchy functional by
\begin{align*}
    g(x + y) &= \log(f(x + y)) = \log(f(x)\cdot f(y))\\ &= \log(f(x))+\log(f(y))= g(x) + g(y).
\end{align*}
Therefore, $g(x)$ must be of the form $\beta\cdot x$.
Hence, 
$f(y) = \exp(\beta\cdot y)$.

\end{proof}

\subsection{Power Function of Circuits}
\label{appx:power}

\begin{theorem}[Hardness of reciprocal of a circuit]
\label{thm:inv}
Let $\p$ be a smooth and decomposable circuit over variables $\X$. 
Then computing $\proj{\p^{-1}(\X)}{\supp(\p)}$ as a decomposable circuit is \#P-Hard, even if $\p$ is structured-decomposable.
\end{theorem}

\begin{proof}
We prove it for the case of PCs over discrete variables. We will prove hardness of computing the reciprocal by showing hardness of computing the partition of the reciprocal of a circuit.
In particular, let $\X=\{X_{1}, \ldots, X_{n}\}$ be a collection of binary variables and let $p$ be a smooth and decomposable PC over $\X$, then computing the quantity
\begin{equation}
    \sum_{\x\in\val(\X)} \frac{1}{p(\x)} \tag{\textsf{INVPC}}
\end{equation}
is \#P-Hard.

Proof is by reduction from the \textsf{EXPLR} problem as defined in \cref{thm:hardness-ratio-pcs}.
Similarly to \cref{thm:hardness-ratio-pcs}, the reduction is built by constructing a smooth and decomposable unnormalized circuit $p(x) = {2^{n}}\cdot{1} + 2^{n}{e^{-(w_0 + \sum_i w_i x_i)}}$. 
The circuit $p$ comprises a sum unit over two sub-circuits.
The first is a uniform (unnormalized) distribution over $\X$ defined as a product unit over $n$ univariate input distribution units that always output 1 for all values $\val(X_i)$ (see \cref{appx:unif-pc} for a construction algorithm).
The second is an exponential of a linear circuit~(\cref{alg:exponential})
and encodes ${e^{-(w_0 + \sum_i w_i x_i)}}$ via a product unit over $n$ univariate input distributions, where one of them encodes $e^{-w_0-w_1x_1}$ and the rest $e^{-w_j x_j}$ for $j=2,\ldots,n$.
Both sub-circuits participates in the sum with parameters $2^{n}$. 

{The size of the constructed circuit is linear in $n$, and \textsf{INVPC} of this circuit corresponds to the solution of the \textsf{EXPLR} problem. 
If you can represent the reciprocal of this circuit as a decomposable circuit, you can compute its marginals (including the partition function) which would solve \textsf{INVPC} and hence  \textsf{EXPLR}.}
Furthermore, the circuit is also omni-compatible because mixture of fully-factorized distributions.
\end{proof}

\begin{theorem}[Hardness of natural power of a decomposable circuit] 
\label{thm:hardness-natural-pow}
Let $\p$ be a smooth and decomposable circuit over variables $\X$. Then computing $\p^{\alpha} (\X)$, for a certain $\alpha\in\N$ as a decomposable circuit is \#P-Hard.
\end{theorem}

\begin{proof}
We prove it for the special case of discrete variables, and by showing the hardness of computing the partition function of $\p^2 (\X)$. In particular, let $\X$ be a collection of binary variables and let $p$ be a smooth and decomposable circuit over $\X$, then computing the quantity
\begin{align*}
    \sum_{\x \in \val(\X)} \p^2 (\x) \tag{\textsf{POW2PC}}
\end{align*}
is \#P-Hard.

The proof builds a reduction from the \textsf{\#3SAT} problem, which is known to be \#P-hard. 
We employ the same setting of \cref{appx:3SAT-circuit}, where a CNF over $n$ Boolean variables $\X = \{X_1, \dots, X_n\}$ and containing $m$ clauses $\{c_1, \dots, c_m\}$, each with exactly 3 literals,   is encoded into two structured-decomposable and deterministic circuits $\p_\beta$ and $\p_\gamma$ over variables $\hat{\X} = \{X_{11}, \dots, X_{1m}, \ldots, X_{n1}, \ldots, X_{nm}\}$.

Then, we construct circuit $\p_\alpha$ as the sum of $\p_\beta$ and $\p_\gamma$,
i.e., $\p_\alpha(\hat{\x}) := \p_\beta(\hat{\x}) + \p_\gamma(\hat{\x})$. 
By definition $\p_\alpha$ is smooth and decomposable, but not structured-decomposable. 
We proceed to show that if we can represent $ \p^2_\alpha(\hat{\x})$ as a smooth and decomposable circuit in polytime, we could solve \textsf{POW2PC} and hence \textsf{\#3SAT}.
That would mean that computing \textsf{POW2PC} is \#P-Hard.

By definition, $\p^2_\alpha (\hat{\x}) = (\p_\beta (\hat{\x}) + \p_\gamma (\hat{\x}))^2 = \p^2_\beta (\hat{\x}) + \p^2_\gamma (\hat{\x}) + 2 \p_\beta (\hat{\x}) \cdot \p_\gamma (\hat{\x})$, and hence 
\begin{align*}
    \sum_{\hat{\x}\in\val(\hat{\X})} \p^2_\alpha (\hat{\x}) & = \sum_{\hat{\x}\in\val(\hat{\X})} \p^2_\beta (\hat{\x}) + \sum_{\hat{\x}\in\val(\hat{\X})} \p^2_\gamma (\hat{\x}) \\
    & + \sum_{\hat{\x}\in\val(\hat{\X})} \p_\beta (\hat{\x}) \cdot \p_\gamma (\hat{\x}).
\end{align*}
Since $\p_\beta$ and $\p_\gamma$ are both structured-decomposable and deterministic the first two summations over the squared circuits can be computed in time $\bigO(\abs{\p_\beta} + \abs{\p_\gamma})$ (see \cref{thm:real-pow-det}). 
It follows that if we could efficiently solve \textsf{POW2PC} we could then solve the that third summation, \ie $\sum_{\hat{\x}\in\val(\hat{\X})} \p_\beta (\hat{\x}) \cdot \p_\gamma (\hat{\x})$.
However, since such a summation is the instance of \textsf{MULPC} between $\p_\beta$ and $\p_\gamma$ 
reduced from \textsf{\#3SAT} (see \cref{thm:prod-dec}), we could solve the latter.
We can conclude that computing \textsf{POW2PC} is \#P-Hard.
\end{proof}

\begin{theorem}[Hardness of natural power of a structured-decomposable circuit]
\label{thm:hardness-natural-pow-sd}
Let $\p$ be a structured-decomposable circuit over variables $\X$. Let $k$ be a natural number. Then there is no polynomial $f(x,y)$ such that the power $\p^k$ can be computed in $\bigO(f(\abs{\p}, k))$ time unless P=NP.
\end{theorem}

\begin{proof}
We construct the proof by showing that for a structured-decomposable circuit $\p$, 
if we could compute
\begin{align*}
    \sum_{\x\in\val(\X)} \p^k (\x).
    \tag{\textsf{POWkPC}}
\end{align*}
in $\bigO(f(\abs{\p}, k))$ time, then we could solve the \textsf{3SAT} problem in polytime, which is known to be NP-Hard.

The \textsf{3SAT} problem is defined as follows: given a set of $n$ Boolean variables $\X = \{X_1, \dots, X_n\}$ and a CNF that contains $m$ clauses $\{c_1, \dots, c_m\}$, each one containing exactly 3 literals, determine whether there exists a satisfiable configuration in $\val(\X)$.

We start by constructing $m$ gadget circuits $\{d_j\}_{j=1}^{m}$ for the $m$ clauses such that $d_j (\x)$ evaluates to $\frac{1}{m}$ iff $\x$ satisfies $c_j$ and otherwise evaluates to $0$, respectively.

Since each clause $c_j$ contains exactly 3 literals, it comprises exactly 7 models w.r.t. the variables appearing in it, \ie its scope $\phi(c_j)$. 
Therefore, following a similar construction in \cref{appx:3SAT-circuit}, we can compile $d_j$ as a weighted sum of 7 circuits that represent the 7 models of $c_j$, respectively. By choosing all weights of $d_j$ as $\frac{1}{m}$, the circuit $d_j$ outputs $\frac{1}{m}$ iff $c_j$ is satisfied; otherwise it outputs $0$.

The gadget circuits $\{d_j\}_{j=1}^{m}$ are then summed together to represent a circuit $\p$. 
That is, $\p = \sumUnit(\{d_j\}_{j=1}^{m}, \{1\}_{j=1}^{m})$. 
In the following, we complete the proof by showing that if the power circuit $\p^k$ 
(we will pick later $k = \ceil{\max(m, n)^2 \cdot \log 2}$) 
can be computed in $\bigO(f(\abs{\p}, k))$ time, then the corresponding \textsf{3SAT} problem can be solved in $\bigO(f(\abs{\p}, k))$ time.

If the original CNF is satisfiable, then there exists at least 1 world such that all clauses are satisfied. In this case, all circuits in $\{d_j\}_{j=1}^{m}$ will evaluate $\frac{1}{m}$. Since $\p$ is the sum of the circuits $\{d_j\}_{j=1}^{m}$, it will evaluate $1$ for any world that satisfies the CNF. 
We obtain the bound
\begin{align*}
    \sum_{\x\in\val(\X)} \p^k (\x) > m \cdot \frac{1}{m} = 1.
\end{align*}

In contrast, if the CNF is unsatisfiable, each variable assignment $\x \in \val(\X)$ satisfies at most $m-1$ clauses, so the circuit $\p$ will output at most $\frac{m-1}{m}$.
Therefore , we retrieve the following bound
\begin{align*}
    \sum_{\x\in\val(\X)} \p^k (\x) \leq 2^{n} \left ( \frac{m-1}{m} \right )^{k}.
\end{align*}

Then, we can retrieve a value for $k$ to separate the two bounds as follows. 
    \begin{align*}
        2^{n} \left ( \frac{m-1}{m} \right )^{k} < 1 \; \Leftrightarrow \; & k > \frac{\log (2^{-n})}{\log \frac{m-1}{m}} \\
        \Leftrightarrow \; & k > \frac{n \log 2}{\log (m) - \log (m-1)} \\
        \overset{(a)}{\Leftrightarrow} \; & k > m \cdot n \cdot \log 2,
    \end{align*}
\noindent where $(a)$ follows the fact that $\log \big ( \frac{m}{m-1} \big ) \leq \frac{1}{m-1}$.
Let $l = \max (m, n)$.
If we choose $k = \ceil{l^2 \cdot \log 2}$, then we can separate the two bounds above.

Therefore, if there exists a polynomial $f(x,y)$ such that the power 
$\p^k$ ($k = \ceil{l^2 \cdot \log 2}$)
can be computed in $\bigO(f(\abs{\p},k))$ time, then we can solve \textsf{3SAT} in $\bigO(f(\abs{\p},k))$ time since the CNF is satisfiable iff $\sum_{\x\in\val(\X)} \p^{k}(\x) > 1$, which is impossible unless P=NP.
\end{proof}

\begin{theorem}[Tractable real power of a deterministic circuit]
\label{thm:real-pow-det}
Let $\p$ be a smooth, decomposable, and deterministic circuit over variables $\X$.
Then, for any real number $\alpha\in\R$, its restricted power, defined as $\proj{a(\x)}{\supp(\p)} = \p^\alpha (\x)\id{\x\in\supp(\p)}$ 
can be represented as a smooth, decomposable, and deterministic circuit over variables $\X$ in $\bigO(\abs{\p})$ time and space.
Moreover, if $\p$ is structured-decomposable, then $a$ is structured-decomposable as well.
\end{theorem}

\begin{proof}

The proof proceeds by construction and recursively builds $\proj{a(\x)}{\supp(\p)}$.
As the base case, we can assume to compute the restricted $\alpha$-power of the input units of $\p$ and represent it as a single new unit.
When we encounter a deterministic sum unit, the power will decompose into the sum of the powers of its inputs. Specifically, let $\p$ be a sum unit: $\p(\X)=\sum_{i\in \ch(p)} \theta_{i} \p_i (\X)$. Then, its  restricted real power circuit $\proj{a(\x)}{\supp(\p)}$ can be expressed as
\begin{align*}
    \proj{a(\x)}{\supp(\p)} &= \left( \sum_{i\in \ch(p)} \theta_{i} \p_i (\x) \right)^{\alpha}\id{\x\in\supp(\p)} \\
    &= \sum_{i\in \ch(p)} \theta_{i}^{\alpha} \big ( \p_i (\x) \big )^{\alpha}\id{\x\in\supp(\p_i)}.
\end{align*}
Note that this construction is possible because only one input of $\p$ is going to be non-zero for any input (determinism).
As such, the power circuit is retaining the same structure of the original sum unit.

Next, for a decomposable product unit, its power will be the product of the powers of its inputs. Specifically, let $\p$ be a product unit: $\p(\X) = \p_1 (\X_1) \cdot p_2 (\X_2)$. Then, its  restricted real power circuit $\proj{a(\x)}{\supp(\p)}$ can be expressed as
\begin{align*}
    &\proj{a(\x)}{\supp(\p)}  = \big ( \p_1 (\x_1) \cdot p_2 (\x_2) \big )^{\alpha} \id{\x\in\supp(\p)}\\
    & = \big ( \p_1 (\x_1) \big )^{\alpha}\id{\x\in\supp(\p_1)} \cdot \big ( \p_2 (\x_2) \big )^{\alpha}\id{\x\in\supp(\p_2)}.
\end{align*}
Note that even this construction preserves the structure of $\p$ and hence its scope partitioning is retained throughout the whole algorithm.
Hence, if $\p$ were also structured-decomposable, then $a$ would be structured-decomposable.
\cref{alg:power-det} illustrates the whole algorithm in detail.

\end{proof}

\begin{algorithm}[tb]
   \caption{\textsc{power}($\p, \alpha, \mathsf{cache}$)}
   \label{alg:power-det}
   \begin{algorithmic}[1]
   \STATE {\bfseries Input:} a smooth, deterministic and decomposable circuit $\p(\X)$, a scalar $\alpha\in\R$, and a cache for memoization 
   \STATE {\bfseries Output:} a smooth, deterministic and decomposable  circuit $a(\X)$ encoding $\proj{\p^\alpha(\X)}{\supp(\p)}$
   \LineIf{$\p\in\mathsf{cache}$}{\textbf{return} $\mathsf{cache}(\p)$}
   \IF{$\p$ is an input unit}
   \STATE $a\leftarrow\inputUnit(\proj{\p^\alpha(\X)}{\supp(\p)}, \scope(p))$
   \ELSIF{$\p$ is a sum unit}
   \STATE $a\leftarrow \sumUnit(\{\text{\textsc{power}}(\p_i, \alpha, \mathsf{cache})\}_{i=1}^{|\ch(\p)|}), \{\theta_i^{\alpha}\}_{i=1}^{|\ch(\p)|})$
   \ELSIF{$\p$ is a product unit}
   \STATE $a\leftarrow \prodUnit(\{\text{\textsc{power}}(\p_i, \alpha, , \mathsf{cache})\}_{i=1}^{|\ch(\p)|})$
   \ENDIF
   \STATE $\mathsf{cache}(\p)\leftarrow a$
   \STATE \textbf{return} $a$
\end{algorithmic}
\end{algorithm}

\begin{theorem}[Tractable natural power of a structured-decomposable circuit]
\label{thm:trac-natural-pow}
Let $\p$ be a structured-decomposable circuit over variables $\X$. Then, for any natural number $\alpha \in \N$, its power circuit $\p^\alpha (\X)$ can be represented as a structured-decomposable circuit over $\X$ in $\bigO(\abs{\p}^\alpha)$ time and space.
\end{theorem}

\begin{proof}
Since $\p$ is compatible with itself, we can run the product algorithm specified in \cref{thm:prod-pcs} recursively to obtain the circuit $\p^\alpha$. By induction, for any $\alpha \in \N$, the size of $\p^\alpha$ is $\bigO(\abs{\p}^\alpha)$.
\end{proof}

\vspace{1em}

\begin{algorithm}[tb]
   \caption{\textsc{power}($\p, \alpha, \mathsf{cache}$) 
   }
   \label{alg:int-power}
   \begin{algorithmic}[1]
   \STATE {\bfseries Input:} a smooth and decomposable circuit $\p(\X)$ and a natural number $\alpha\in\N$.
   \STATE {\bfseries Output:} a smooth and decomposable circuit $a$ over $\X$ encoding $a (\X)=(\p(\X))^{\alpha}$.
   \STATE $ a \leftarrow p$
   \STATE $r \leftarrow \mod(\alpha, 2)$
   \WHILE{$\alpha > 1$}
    \STATE $a \leftarrow \text{\textsc{multiply}}(a, a)$
    \STATE $\alpha \leftarrow \floor{\alpha / 2}$
   \ENDWHILE
   \IF{$r = 1$}
   \STATE $a \leftarrow \text{\textsc{multiply}}(a, p);$ 
   \ENDIF
   \STATE \textbf{return} $a$
\end{algorithmic}
\end{algorithm}

\subsection{Quotient of Circuits}
\label{appx:quotient}

\begin{theorem}[Hardness of quotient of two circuits]
\label{thm:hardness-ratio-pcs}
Let $\p$ and $\q$ be two smooth and decomposable circuits over variables $\X$, and let $\q(\x)\neq 0$ for every $\x\in\val(\X)$.
Then, computing their quotient $\p(\X)/\q(\X)$
as a decomposable circuit is \#P-Hard, even if they are compatible.
\end{theorem}

\begin{proof}
This result follows from \cref{thm:inv} by noting that computing the reciprocal of a circuit is a special case of computing the quotient of two circuits. In particular, let $\p$ be an omni-compatible circuit representing the constant function $1$ over variables $\X$, constructed as in \cref{appx:unif-pc}.
Then computing the reciprocal of a structured-decomposable circuit $\q$ as a decomposable circuit reduces to computing the quotient $\p/\q$.
\end{proof}

\begin{theorem}[Tractable restricted quotient of two circuits]
\label{thm:ratio-pcs}
Let $\p$ and $\q$ be two compatible circuits over variables $\X$, and let $\q$ be also deterministic. %
Then, their quotient restricted to $\supp(q)$ %
can be represented as a circuit compatible with $\p$ (and $\q$) over variables $\X$  in time and space $\bigO(\abs{\p}\abs{\q})$.
Moreover, if $\p$ is also deterministic, then the quotient circuit is deterministic as well.
\end{theorem}

\begin{proof}
We know from \cref{thm:real-pow-det} that we can obtain the reciprocal circuit $\q^{-1}$ that is also compatible with $\q$ (and by extension $\p$) in $\bigO(\abs{\q})$ time and space.
Then we can multiply $\p$ and $\q^{-1}$ in $\bigO(\abs{\p}\abs{\q})$ time using \cref{thm:prod-pcs} to compute their quotient circuit that is still compatible with $\p$ and $\q$.
If $\p$ is also deterministic, then we are multiplying two deterministic circuits and therefore  their product circuit is deterministic (\cref{thm:prod-pcs}).
\end{proof}

\subsection{Logarithm of a PC}

\begin{theorem}[Hardness of the logarithm of a circuit]
\label{thm:hardness-log}
Let $\p$ be a smooth and decomposable PC over variables $\X$.
Then, computing its logarithm circuit $l(\X) := \log \p(\X)$ as a decomposable circuit is \#P-Hard, even if $\p$ is structured-decomposable.
\end{theorem}

\begin{proof}
We will prove hardness of computing the logarithm by showing hardness of computing the partition function of the logarithm of a circuit. Let $\X=\{X_{1}, \ldots, X_{n}\}$ be a collection of binary variables, and $p$ a smooth and decomposable PC over $\X$ where $p(\x)>0$ for all $\x\in\val(\X)$. Then computing the quantity
\begin{equation}
    \sum_{\x\in\val(\X)} \log p(\x) \tag{\textsf{LOGPC}}
\end{equation}
is \#P-Hard.

The proof is by reduction from \textsf{\#NUMPAR}, the counting problem of the number partitioning problem (\textsf{NUMPAR}) defined as follows. Given $n$ positive integers $k_1,\dots,k_n$, we want to decide whether there exists a subset $S\subset[n]$ such that $\sum_{i\in S} k_i = \sum_{i\not\in S} k_i$. \textsf{NUMPAR} is NP-complete, and \textsf{\#NUMPAR} which asks for the number of solutions is known to be \#P-hard.

We will show that we can solve \textsf{\#NUMPAR} using an oracle for \textsf{LOGPC}, which will imply that \textsf{LOGPC} is also \#P-hard.
First, consider the following quantity $\mathsf{SL}$ for a given weight function $w(\cdot)$:
\begin{align*}
    \mathsf{SL} &:= \sum_{\x\in\val(\X)} \log(\sigma(w(\x))+1) \\
    &= \sum_{\x\in\val(\X)} \log\left(\frac{1}{1+e^{-w(\x)}}+1\right) \\
    &= \sum_{\x\in\val(\X)} \log\left(\frac{2+e^{-w(\x)}}{1+e^{-w(\x)}}\right) \\
    &= \sum_{\x\in\val(\X)}\!\! \log(2+e^{-w(\x)}) - \!\!\!\! \sum_{\x\in\val(\X)}\!\! \log(1+e^{-w(\x)}).
\end{align*}
Similar to the construction in the proof of \cref{thm:inv},
we can construct smooth and decomposable, unnormalized PCs for $2+e^{-w(\x)}$ and $1+e^{-w(\x)}$ of size linear in $n$. Then, we can compute $\mathsf{SL}$ via two calls to the oracle for $\textsf{LOGPC}$ on these PCs.

Next, we choose the weight function $w(\cdot)$ such that $\mathsf{SL}$ can be used to answer \textsf{\#NUMPAR}. For a given instance of \textsf{NUMPAR} described by $k_1,\dots,k_n$ and a large integer $m$, which will be chosen later, we define the following weight function:
\begin{equation*}
    w(\x) := -\frac{m}{2} - m \sum_i k_i + 2 m \sum_i k_i x_i.
\end{equation*}
In other words, $w(\x)=w_0+\sum_i w_i x_i$ where $w_0=-m/2-m\sum_i k_i$ and $w_i=2mk_i$ for $i=1,\dots,n$.
Here, an assignment $\x$ corresponds to a subset $S_{\x} = \{i | x_i=1, x_i\in\x \}$. Then the assignment $1-\x$ corresponds to the complement $S_{1-\x}=\overline{S_{\x}}$.
In the following, we will consider pairs of assignments $(\x,1-\x)$ and say that it is a solution to \textsf{NUMPAR} if $S_{\x}$ and by extension $S_{1-\x}$ are solutions to \textsf{NUMPAR}.

Observe that if $(\x,1-\x)$ is a solution to \textsf{NUMPAR}, then $w(\x)=w(1-\x)=-m/2$. Otherwise, one of their weights must be $\geq m/2$ and the other $\leq -3m/2$.
We can then deduce the following facts about the \textit{contribution} of each pair to $\mathsf{SL}$, defined as $c(\x,1-\x)=\log(\sigma(w(\x))+1)+\log(\sigma(w(1-\x))+1)$.

If the pair $(\x,1-\x)$ is a solution to \textsf{NUMPAR}, then its contribution to \textsf{SL} is going to be:
\begin{align*}
    c(\x,1-\x) = 2\log(\sigma(-m/2)+1).
\end{align*}
Otherwise, we can bound its contribution as follows:
\begin{align*}
    \log(\sigma(m/2)+1) \leq c(\x,1-\x) \leq 1 + \log(\sigma(-3m/2)+1) 
\end{align*}

If there are $k$ pairs that are solutions to the \textsf{NUMPAR} problem, then using the above observations we have the following bounds on \textsf{SL}:
\begin{align}
    \mathsf{SL} 
    \geq &(2^{n-1}-k) \log\left(\sigma(m/2)+1\right) \nonumber\\
    &+ 2k \log\left(\sigma(-m/2)+1\right) \nonumber\\
    \geq &(2^{n-1}-k) \log\left(\sigma(m/2)+1\right), \label{eq:sl-lb}
\end{align}
\begin{align}
    \mathsf{SL}
    \leq &(2^{n-1}-k) (1+\log\left(\sigma(-3m/2)+1\right)) \nonumber\\
    &+ 2k \log(\sigma(-m/2)+1). \label{eq:sl-ub}
\end{align}

Suppose for some given $\epsilon>0$, we select $m$ such that it satisfies both $1-\epsilon \leq \log(\sigma(m/2)+1)$ and $\log(\sigma(-m/2)+1) \leq \epsilon$.
First, this implies that $m$ also satisfies the following: 
\begin{align*}
    &1+\log\left(\sigma(-3m/2)+1)\right) \\
    &\leq 1 + \log(\sigma(-m/2)+1)
    \leq 1 + \epsilon.
\end{align*}

Plugging in above inequalities to \cref{eq:sl-lb,eq:sl-ub}, we get the following bounds on \textsf{SL} in terms of $\epsilon$ and $k$:
\begin{align*}
    (2^{n-1} -k)(1-\epsilon) 
    \leq \mathsf{SL}
    \leq (2^{n-1}-k)(1+\epsilon)+2k\epsilon.
\end{align*}
We can alternatively express this as the following bounds on $k$: 
\begin{align*}
    \frac{2^{n-1}(1-\epsilon) - \mathsf{SL}}{1-\epsilon}
    \leq k 
    \leq \frac{2^{n-1}(1+\epsilon) - \mathsf{SL}}{1-\epsilon}.
\end{align*}
The difference between the upper and lower bounds on $k$ is equal to $2^n \epsilon / (1-\epsilon)$.
If this difference is less than 1---for example by setting $\epsilon = 1/(2^n+2)$---we can exactly solve for $k$. In particular, it must be equal to the ceiling of the lower bound as well as the floor of the upper bound.
Moreover, the answer to \textsf{\#NUMPAR} is given by $2k$. This concludes the proof that computing \textsf{LOGPC} is \#P-hard.
\end{proof}

\begin{theorem}[Tractable logarithm of a circuit]
\label{thm:log-det}
Let $\p$ be a smooth, deterministic and decomposable PC over variables $\X$. Then its logarithm circuit, restricted to the support of $\p$ and defined as 
\begin{equation*}
    \proj{l(\x)}{\supp(\p)} = \begin{cases}
    {\log\p(\x)} &\text{if $\x\in\supp(\p)$}\\
    0 &\text{otherwise}
    \end{cases}
\end{equation*}
for every $\x\in\val(\X)$ can be represented as a smooth and decomposable circuit that shares the scope partitioning of $\p$ in $\bigO(\abs{\p})$ time and space.
\end{theorem}

\begin{proof}

The proof proceeds by recursively constructing $\proj{l(\x)}{\supp(\p)}$. In the base case, we assume computing the logarithm of an input unit can be done in $\bigO(1)$ time. 
When we encounter a deterministic sum unit $\p(\x)=\sum_{i \in |\ch(\p)|} \theta_i \p_i (\x)$, its logarithm circuit consists of the sum of (i) the logarithm circuits of its child units and (ii) the support circuits of its children weighted by their respective weights $\{\theta_i\}_{i=1}^{\abs{\ch(\p)}}$: 
\begin{align*}
    & \proj{l(\x)}{\supp(\x)} = \log \left (\sum_{i \in \ch(\p)} \theta_i \p_i (\x) \right )\cdot\id{\x\in\supp(\p)} \\
    & = \sum_{i \in |\ch(\p)|} \log \Big (\theta_i \p_i (\x) \Big ) \id{\x\in\supp(p_i)} \\
    & = \sum_{i \in |\ch(\p)|} \log \theta_i \id{\x\in\supp(p_i)} + \sum_{i \in |\ch(\p)|} \proj{l_i (\x)}{\supp(\p_i)}.
\end{align*}

For a smooth, decomposable, and deterministic product unit $\p(\x) = \p_1 (\x) \p_2 (\x)$, its logarithm circuit can be decomposed as sum of the logarithm circuits of its child units:
\begin{align*}
    & \proj{l(\x)}{\supp(\x)} = \log \left ( \p_1 (\x_1) \p_2 (\x_2) \right )\cdot\id{\x\in\supp(\p)} \\
    & = \log \p_1 (\x_1) \id{\x\in\supp(\p)} + \log \p_2 (\x_2) \id{\x\in\supp(\p)} \\
    & = \log \p_1 (\x_1)\id{\x_1\in\supp(\p_1)} \id{\x_2\in\supp(\p_2)} + \\
    & \log \p_2 (\x_2) \id{\x_2\in\supp(\p_2)}\id{\x_1\in\supp(\p_1)} =\\
    &\proj{l(\x_1)}{\supp(\p_1)}\id{\x_2\in\supp(\p_2)} + \\ &\proj{l(\x_2)}{\supp(\p_2)}\id{\x_1\in\supp(\p_1)}.
\end{align*}

Note that in both case, the support circuits (e.g., $\id{\x\in\supp(\p)}$) are used to enforce smoothness in the output circuit. \cref{alg:log} illustrates the whole algorithm in detail,
showing that the construction of these support circuits can be done in linear time by caching intermediate sub-circuits while calling \cref{alg:support-det}. 
Furthermore, the newly introduced product units, \ie $\proj{l(\x_1)}{\supp(\p_1)}\id{\x_2\in\supp(\p_2)}$, $\proj{l(\x_2)}{\supp(\p_2)}\id{\x_1\in\supp(\p_1)}$, and the additional support input unit $\log \theta_i \id{\x\in\supp(p_i)}$ share the same support of $\p$ by construction.
This implies that when a deterministic circuit and its logarithmic circuit are going to be multiplied, \eg when computing entropies (\cref{appx:entropies}), we can check for their support to overlap in linear time (\cref{alg:prod}).

\end{proof}

\begin{algorithm}[tb]
   \caption{\textsc{logarithm}($\p, \mathsf{cache}_l, \mathsf{cache}_s$)}
   \label{alg:log}
   \begin{algorithmic}[1]
   \STATE {\bfseries Input:} a smooth, deterministic and decomposable PC $\p(\X)$ and two caches for memoization ($\mathsf{cache}_l$ for the logarithmic circuit and $\mathsf{cache}_s$ for the support circuit).
   \STATE {\bfseries Output:} a smooth
   and decomposable circuit $l(\X)$ encoding $\log\left(\p(\X)\right)$
    \LineIf{$\p \in \mathsf{cache}_l$}{\textbf{return} $\mathsf{cache}_l(\p)$}
   \IF{$\p$ is an input unit}
   \STATE $l\leftarrow\inputUnit(\log\left(\p_{\mid\supp(\p)}\right), \scope(p))$
   \ELSIF{$\p$ is a sum unit}
   \STATE $\node\leftarrow\{\}$
   \FOR{$i=1$ \textbf{to} $|\ch(p)|$}
   \STATE $\node\leftarrow \node\cup\{\suppCirc(p_i, \mathsf{cache}_s)\}$
   \STATE $\node\leftarrow \node\cup\{\text{\textsc{logarithm}}(p_i, \mathsf{cache}_l)\}$
   \ENDFOR
   \STATE $l\leftarrow\sumUnit(\node, \{\log\theta_1, 1, \log\theta_2, 1, \ldots, \log\theta_{|\ch(\p)|}, 1\})$
   \ELSIF{$\p$ is a product unit}
   \STATE $n \leftarrow \{\}$
   \FOR{$i=1$ \textbf{to} $|\ch(p)|$}
     \STATE $n \leftarrow n \cup \{\prodUnit(\{\text{\textsc{logarithm}}(p_i, \mathsf{cache}_l)\}\cup\{\suppCirc(p_j, \mathsf{cache}_s)\}_{j\neq i})\}$
   \ENDFOR
   \STATE $l \leftarrow \sumUnit(n, \{1\}_{i=1}^{|\ch(\p)|})$
   \ENDIF
   \STATE $\mathsf{cache}_l(\p)\leftarrow l$
   \STATE \textbf{return} $l$
\end{algorithmic}
\end{algorithm}

\subsection{Exponential Function of a Circuit}

\begin{theorem}[Hardness of the exponential of a circuit]
\label{thm:hardness-exp}
Let $\p$ be a smooth and decomposable circuit over variables $\X$.
Then, computing its exponential $\exp\left(\p(\X)\right)$ as a decomposable circuit is \#P-Hard, even if $\p$ is structured-decomposable. 
\end{theorem}

\begin{proof}
We will prove hardness of computing the exponential by showing hardness of computing the partition function of the exponential of a circuit. Let $\X=\{X_{1}, \ldots, X_{n}\}$ be a collection of binary variables with values in $\{-1, +1\} $ and let $p$  be a smooth and decomposable PC over $\X$  then computing the quantity
\begin{equation}
    \sum_{\x\in\val(\X)} \exp\left(\p(\x)\right) \tag{\textsf{EXPOPC}}
\end{equation}
is \#P-Hard.

The proof is a reduction from the problem of computing the partition function of an Ising model, \textsf{ISING} which is known to be \#P-complete~\citep{jerrum1993polynomial}.
Given a graph $G=(V, E)$ with $n$ vertexes, computing the partition function of an Ising model associated to $G$ and equipped with potentials associated to its edges ($\{w_{u,v}\}_{(u,v)\in E}$) and vertexes ($\{w_v\}_{v\in V}$) equals to
\begin{equation}
    \sum_{\x\in\val(\X)} \exp\left(\sum_{(u,v)\in E}w_{u,v}x_u x_v + \sum_{v\in V}w_v x_v\right). \tag{\textsf{ISING}}
\end{equation}

The reduction is made by constructing a smooth and decomposable circuit $\p(\X)$ that computes $\sum_{(u,v)\in E}w_{u,v}x_u x_v + \sum_{v\in V}$.
This can be done by introducing a sum units with $\abs{E} + \abs{V}$ inputs that are product units and with weights $\{w_{u,v}\}_{(u,v)\in E}\cup\{w_v\}_{v\in V}$.
The first $\abs{E}$ product units receive inputs from $n$ input distributions where only 2 corresponds to the binary indicator inputs $X_u$ and $X_v$ for an edge $(u,v)\in E$ while the remaining $n-2$ are uniform distributions outputting 1 for all the possible states of variables $\X\setminus\{X_u, X_v\}$.
Analogously, the remaining $\abs{V}$ product units receive input from $n$ of which only one, corresponding to the vertex $v \in V$ is an indicator unit over $X_v$, while the remaining are uniform distributions for variables in $\X\setminus\{X_v\}$.
\end{proof}

\begin{proposition}[Tractable exponential of a linear circuit]
\label{thm:exp-linear}
Let $\p$ be a linear circuit over variables $\X$, \ie $\p(\X)=\sum_{i}\theta_i\cdot X_i$. Then $\exp\left(\p(\X)\right)$ can be represented as an omni-compatible circuit with a single product unit in $\bigO(\abs{\p})$ time and space.
\begin{proof}
The proof follows immediately by the properties of exponentials of sums.
\cref{alg:exponential} formalizes the construction.
\end{proof}
\end{proposition}

\begin{algorithm}[H]
   \caption{\textsc{exponential}($\p$)}
   \label{alg:exponential}
   \begin{algorithmic}[1]
   \STATE {\bfseries Input:} a smooth circuit $\p$ over variables $\X=\{X_1,X_2,\ldots,X_n\}$ encoding $\p(\X)=\theta_0+\sum_{i=1}^{n}\theta_i X_i$
   \STATE {\bfseries Output:} its exponential circuit encoding $\exp\left(\p(\X)\right)$
   \STATE $e\leftarrow \{\inputUnit(\exp\left(\theta_0+\theta_1 X_1\right), X_1)\}$
   \FOR {$i=2$ \textbf{to} $n$}
   \STATE $e\leftarrow e\cup\{\inputUnit(\exp\left(\theta_i X_i\right), X_i)\}$
   \ENDFOR
   \STATE \textbf{return} $\prodUnit(e)$
\end{algorithmic}
\end{algorithm}

\section{Information-Theoretic Queries}
\label{appx:info-theory}

\subsection{Cross Entropy}

\begin{theorem}[Hardness of cross-entropy of two PCs]
\label{thm:hardness-cross-entropy}
Let $\p$ and $\q$ be two smooth and decomposable PCs over variables $\X$.
Then, computing their cross-entropy, \ie
\begin{equation*}
    \label{eq:cross-entropy}
    -\int_{\val(\X)} \p(\x)\log(\q(\x))d\X
\end{equation*}
 is \#P-Hard, even if $\p$ and $\q$ are compatible over $\X$.
\end{theorem}

\begin{proof}
The proof consists of a simple reduction from \textsf{LOGPC} from \cref{thm:hardness-log}. We know that computing \textsf{LOGPC} for a smooth and decomposable PC over binary variables $\X$ is \#P-hard. 
We can reduce this to computing the cross entropy between $\p=1$, which can be constructed as an omni-compatible circuit (\cref{appx:unif-pc}), and the original PC of the \textsf{LOGPC} problem. Thus, the cross-entropy of two compatible circuits is a \#P-hard problem.
\end{proof}

\begin{theorem}[Tractable cross-entropy of two PCs]
\label{thm:cross-entropy}
Let $\p$ and $\q$ be two compatible PCs over variables $\X$, and also let $\q$ be deterministic.
Then their cross-entropy restricted to the support of $\q$ can be exactly computed in $\bigO(\abs{\p}\abs{\q})$ time and space.
\end{theorem}

\begin{proof}

From \cref{thm:log-det} we know that we can compute the logarithm of $\q$ in polytime, which is a PC of size $\bigO(\abs{\q})$ that is compatible with $\q$ and hence with $\p$. 
Therefore, multiplying $\p$ and $\log\q$ according to \cref{thm:prod-dec} can be done exactly in polytime and yields a circuit of size $\bigO(\abs{\p}\abs{\q})$ that is still smooth and decomposable, hence we can tractably compute its partition function.
\end{proof}

\subsection{Entropy}
\label{appx:entropies}

\begin{theorem}[Hardness of the Shannon entropy of a PC]
\label{thm:hardness-entropy}
Let $\p$ be a smooth and decomposable PC over variables $\X$.
Then, computing its entropy, defined as
\begin{equation*}
    \label{eq:entropy}
    \text{\textsc{ent}}(p) := 
    -\sum_{\val(\X)} \p(\x)\log(\p(\x))d\X \tag{$\mathsf{ENTPC}$}
\end{equation*}
is coNP-Hard.
\end{theorem}

\begin{proof}
The hardness proof contains a polytime reduction from the coNP-hard \textsf{3UNSAT} problem, defined as follows: given a set of $n$ Boolean variables $\X = \{X_1, \dots, X_n\}$ and a CNF with $m$ clauses $\{c_1, \dots, c_m\}$ (each clause contains exactly 3 literals), decide whether the CNF is unsatisfiable.

The reduction borrows two gadget circuits $\p_\beta$ and $\p_\gamma$ defined in \cref{appx:3SAT-circuit}.
They each represent a logical formula over an auxiliary set of variables, which we denote here $\X^\prime$, and thus outputs 0 or 1 for all values of $\X^\prime$.
Moreover, by construction, $\p_\beta \cdot \p_\gamma$ is the constant function 0 if and only if the original CNF is unsatisfiable.

We further construct a circuit $p_{\alpha}$ as the summation over $p_{\beta}$ and $p_{\gamma}$. 
Recall that $p_\beta$ and $p_\gamma$ can efficiently be constructed as smooth and decomposable circuits, and thus their sum can be represented as a smooth and decomposable circuit in polynomial time.
We will now show that \textsf{3UNSAT} can be reduced to checking whether the entropy of $p_\alpha$ is zero. 
 
First, observe that for any assignment $\x^\prime$ to $\X^\prime$, $p_\alpha(\x^\prime)$ evaluates to 0, 1, or 2, because $p_\beta$ and $p_\gamma$ always evaluates to either 0 or 1.
Moreover, if $p_\alpha$ only outputs 0 or 1 for all values of $\X^\prime$, then $p_\beta \cdot p_\gamma$ must always be 0, implying that the original CNF is unsatisfiable.
Lastly, in such a case, the entropy of $p_\alpha$ must be 0, whereas the entropy will be nonzero if there is an assignment $\x^\prime$ such that $p_\alpha(\x^\prime)=2$. 
This concludes the proof that computing the entropy of a smooth and decomposable PC is coNP-hard.
\end{proof}

\begin{theorem}[Tractable entropy of a PC]
\label{thm:entropy}
Let $\p$ be a smooth, deterministic, and decomposable PC over variables $\X$.
Then its entropy,\footnote{For the continuous case this quantity refers to the \textit{differential entropy}, while for the discrete case it is the Shannon entropy.} defined as 
\begin{equation*}
    -\int_{\val(\X)}\p(\x)\log\p(\x)\:d\X
\end{equation*}
can be exactly computed in $\bigO(\abs{\p})$ time and space.
\end{theorem}

\begin{proof}
From \cref{thm:log-det} we know that we can compute the logarithm of $\p$ in polytime as a smooth and decomposable PC of size $\bigO(\abs{\p})$ which furthermore shares the same support partitioning with $\p$. 
Therefore, multiplying $\p$ and $\log\p$ according to \cref{alg:prod} can be done in polytime and yields a smooth and decomposable circuit of size $\bigO(\abs{\p})$ since $\log\p$ shares the same support structure of $\p$ (\cref{thm:log-det}).
Therefore, we can compute the partition function of the resulting circuit in time linear in its size.
\end{proof}

\subsection{Mutual Information}

\begin{theorem}[Hardness of the mutual information of a PC]
\label{thm:hardness-mi}
Let $p$ be a smooth, decomposable, and deterministic PC over variables $\Z = \X \cup \Y$ ($\X \cap \Y = \emptyset$). Then, computing the mutual information between $\X$ and $\Y$, defined as 
\begin{align*}
    \text{\textsc{mi}}(p; \X, \Y) := \int_{\val(\Z)} p(\x,\y) \log \frac{p(\x,\y)}{p(\x)\cdot p(\y)}d\X d\Y 
\end{align*}
is coNP-Hard.
\end{theorem}

\begin{proof}
We show hardness for the case of Boolean inputs, which implies hardness in the general case.
This proof largely follows the proof of \cref{thm:hardness-entropy} to show that there is a polytime reduction from \textsf{3UNSAT} to the mutual information of PCs.
For a given CNF, suppose we construct $p_\beta$, $p_\gamma$, and $p_\alpha=p_\beta+p_\gamma$ over a set of Boolean variables, say $\X$, as shown in \cref{appx:unif-pc,thm:hardness-entropy}.

Let $\Y\!=\!\{Y\}$ be a single Boolean variable, and define $p_\delta$ as:
\begin{align*}
    p_\delta := p_\beta \times \id{Y=1} + p_\gamma \times \id{Y=0}.
\end{align*}
That is, we first construct two product units $q_1$, $q_2$ with inputs $\{p_\beta, \id{Y=1}\}$ and $\{p_\gamma, \id{Y=0}\}$, respectively, and build a sum unit $p_\delta$ with inputs $\{q_1, q_2\}$ and weights $\{1, 1\}$. Then $p_\delta$ has the following properties:
\textbf{(1)} $p_\delta$ is smooth, decomposable, and deterministic, following from the fact that $p_\beta$ and $p_\gamma$ are also smooth, decomposable, and deterministic, and that $q_1$ and $q_2$ have no overlapping support.
\textbf{(2)} $\text{\textsc{ent}}(p_\delta)$ can be computed in linear-time w.r.t. the circuit size by \cref{thm:entropy}.
\textbf{(3)} $p_\delta (Y = 1)$ and $p_\delta (Y = 0)$ can be computed in linear time (w.r.t. size of the circuit $p_\delta$), as $p_\delta$ admits tractable marginalization.
\textbf{(4)} For any $\x\in\val(\X)$, $p_\delta(\x)=p_\beta(\x)+p_\gamma(\x)=p_\alpha(\x)$.

We can express the mutual information $\text{\textsc{mi}}(p_\delta; \X, \Y)$ as:
\begin{align*}
    \text{\textsc{mi}}(p_\delta; \X, \Y) =& \text{\textsc{ent}}(p_\delta) - p_\delta (Y \!=\! 1) \log p_\delta (Y \!=\! 1) \\
    &\: - p_\delta (Y \!=\! 0) \log p_\delta (Y \!=\! 0) - \text{\textsc{ent}}(p_\alpha).
\end{align*}
Therefore, given an oracle that computes $\text{\textsc{mi}}(p_\delta; \X, \Y)$, we can check if it is equal to $\text{\textsc{ent}}(p_\delta) - p_\delta (Y = 1) \log p_\delta (Y = 1) - p_\delta (Y = 0) \log p_\delta (Y = 0)$, which is equivalent to checking $\text{\textsc{ent}}(p_\alpha)=0$, and decide whether the original CNF is unsatisfiable.
Hence, computing the mutual information of smooth, deterministic, and decomposable PCs is a coNP-hard problem.
\end{proof}

\begin{theorem}[Tractable mutual information of a PCs]
\label{thm:mi}
Let $\p$ be a deterministic and structured-decomposable PC over variables $\Z = \X \cup \Y$ ($\X \cap \Y = \emptyset$).
Then the mutual information between $\X$ and $\Y$
can be exactly computed in $\bigO(\abs{\p})$ time and space if $\p$ is still deterministic after marginalizing out $\Y$ as well as after marginalizing out $\X$.\footnote{This structural property of circuits is also known as marginal determinism~\citep{choi2020pc} and has been introduced in the context of marginal MAP inference and the computation of same-decision probabilities of Bayesian classifiers~\citep{oztok2016solving,choi2017optimal}.}
\end{theorem}

\begin{proof}
From \cref{thm:log-det} we know that the logarithm circuits of $\p(\X,\Y)$, $\p(\X)\id{\y \in \supp(\p(\Y))}$, and $\p(\Y)\id{\x \in \supp(\p(\X))}$ can be computed in polytime and are smooth and decomposable circuits of size $\bigO(\abs{\p})$ that furthermore share the same support partitioning with $\p(\Y,\Z)$.
Therefore, we can multiply $\p(\X,\Y)$ with each of these logarithm circuits efficiently 
according to \cref{thm:prod-pcs} to yield circuits of size $\bigO(\abs{\p})$.
These are still smooth and decomposable circuits. Hence we can compute their partition functions and compute the mutual information between $\X$ and $\Y$ \wrt $\p$.
\end{proof}

\subsection{Divergences}

\subsubsection{Kullback-Leibler Divergence}

\begin{definition}[Kullback-Leibler divergence]
\label{def:kld}
The Kullback-Leibler divergence (KLD)\footnote{Also called intersectional KLD in\citet{LiangXAI17} since the integral is restricted over the intersection of the supports of the two PCs.} of two PCs $\p$ and $\q$ is defined as 
\begin{align*}
    {\mathbb{D}}_{\mathsf{KL}}(\p\parallel\q) = \int_{\supp(\p)\cap\supp(\q)} \p(\x)\log\frac{\p(\x)}{\q(\x)}d\X.
\end{align*}
\end{definition}

\begin{theorem}[Hardness of KLD of two PCs]
\label{thm:hardness-kld}
Let $\p$ and $\q$ be two smooth and decomposable PCs over variables $\X$. 
Then, computing their Kullback-Leibler divergence 
is \#P-Hard,
even if $\p$ and $\q$ are compatible. 
\end{theorem}

\begin{proof}
The proof proceeds similarly to the proof of \cref{thm:hardness-cross-entropy}.
Recall that the \textsf{LOGPC} problem from \cref{thm:hardness-log} is \#P-hard for a smooth and decomposable PC over binary variables. 
We can reduce this to computing the negative of KL divergence between $\p=1$, which can be constructed as an omni-compatible circuit (\cref{appx:unif-pc}), and $\q$ the original PC of the \textsf{LOGPC} problem. Thus, the KLD of two compatible circuits is a \#P-hard problem.
\end{proof}

\begin{theorem}[Tractable KLD of two PCs]
\label{thm:kld}
Let $\p$ and $\q$ be two deterministic and compatible PCs over variables $\X$.
Then, their intersectional KLD can exactly be computed in time and space $\bigO(\abs{\p}\abs{\q})$.
\end{theorem}

\begin{proof}
Tractability of the intersectional KLD can be concluded directly from the tractability of cross entropy and entropy (\cref{thm:cross-entropy,thm:entropy}). Specifically, KLD can be expressed as the difference between cross entropy and entropy:
\begin{align*}
    &\int \p(\x)\log\frac{\p(\x)}{\q(\x)}\: d\X = \int \p(\x)\log{{\p(\x)}}\: d\X - \int \p(\x)\log{{\q(\x)}}\: d\X.
\end{align*}
We can compute the entropy of a smooth, decomposable, and deterministic PC $\p$ in $\bigO(\abs{\p})$; and the cross entropy between two deterministic and compatible PCs $\p$ and $\q$ in $\bigO(\abs{\p}\abs{\q})$ time.
\end{proof}

\subsubsection{R\'enyi Entropy}

\begin{definition}[Rényi entropy]
\label{def:renyi-entropy}
The Rényi entropy of order $\alpha\in\R$ of a PC $\p$ is defined as
\begin{align*}
    \frac{1}{1-\alpha} \log \int_{\supp(\p)} \p^\alpha (\x) d\X.
\end{align*}
\end{definition}

\begin{theorem}[Hardness of Rényi entropy for natural $\alpha$] 
\label{thm:hardness-reiny-entropy-natural}
Let $\p$ be a smooth and decomposable PC over variables $\X$, and $\alpha$ be a natural number. Then computing its Rényi entropy of order $\alpha$
is \#P-Hard.
\end{theorem}

\begin{proof}
We show hardness for the case of discrete inputs.
The hardness of computing the Rényi entropy for natural number $\alpha$ is implied by the hardness of computing the natural power of smooth and decomposable PCs, which is proved in \cref{thm:hardness-natural-pow}. Specifically, we conclude the proof by observing that there exists a polytime reduction from \textsf{POW2PC}, defined as $\sum_{\x\in\val(\X)} \p^2(\x)$, a \#P-Hard problem as proved in \cref{thm:hardness-natural-pow}, to Rényi entropy with $\alpha = 2$.
\end{proof}

\begin{theorem}[Hardness of Rényi entropy for real $\alpha$] 
\label{thm:hardness-reiny-entropy-real}
Let $\p$ be a structured-decomposable PC over variables $\X$ and $\alpha$ be a non-natural real number. Then computing its Rényi entropy of order $\alpha$ is \#P-Hard.
\end{theorem}

\begin{proof}
Similar to the proof of \cref{thm:hardness-reiny-entropy-natural}, this hardness result follows from the fact that computing the reciprocal of a structured-decomposable circuit is \#P-Hard (\cref{thm:inv}). Again, this is demonstrated by a polytime reduction from \textsf{INVPC} (i.e., $\sum_{\x\in\val(\X)} p^{-1}(\x)$) to Rényi entropy with $\alpha = -1$.
\end{proof}

\begin{theorem}[Tractable Rényi entropy for natural $\alpha$]
\label{thm:trac-reiny-natural}
Let $\p$ be a structured-decomposable PC over variables $\X$ and $\alpha \in \N$. Its Rényi entropy can be computed in $\bigO(\abs{\p}^\alpha)$ time.
\end{theorem}
\begin{proof}
The proof easily follows from computing the natural power circuit of $\p$, which takes $\bigO(\abs{\p}^\alpha)$ time according to \cref{thm:trac-natural-pow}.
\end{proof}

\begin{theorem}[Tractable Rényi entropy for real $\alpha$]
\label{thm:trac-reiny-real}
Let $\p$ be a smooth, decomposable, and deterministic PC over variables $\X$ and $\alpha \in \R_{+}$. Its Rényi entropy can be computed in $\bigO(\abs{\p})$ time and space.
\end{theorem}
\begin{proof}
The proof easily follows from computing the power circuit of $\p$, which takes $\bigO(\abs{\p})$ time according to \cref{thm:real-pow-det}.
\end{proof}

\subsubsection{R\'enyi's $\alpha$-divergence}

\begin{definition}[Rényi's $\alpha$-divergence]
\label{def:renyi-div}
The Rényi's $\alpha$-divergence of two PCs $\p$ and $\q$ is defined as
\begin{align*}
    {\mathbb{D}}_{\alpha}(\p\parallel\q) = \frac{1}{1-\alpha}\log\int_{\supp(\p)\cap\supp(\q)}\p^{\alpha}(\x)\q^{1-\alpha}(\x)d\X.
\end{align*}
\end{definition}

\begin{theorem}[Hardness of alpha divergence of two PCs]
\label{thm:hard-alpha-div}
Let $\p$ and $\q$ be two smooth and decomposable PCs over variables $\X$. Then computing their Rényi's $\alpha$-divergence 
for $\alpha\in\R\setminus\{1\}$ is \#P-Hard, even if $\p$ and $\q$ are compatible.
\begin{proof}
Suppose $\p$ is a smooth and decomposable PC $\X$ representing the constant function 1, which can be constructed as in \cref{appx:unif-pc}. Then $\p^\alpha$ is also a constant 1. 
Hence, computing R\'enyi's $2$-divergence between $\p$ and another smooth and decomposable PC $\q$ is as hard as computing the reciprocal of $\q$, which is \#P-hard (\cref{thm:inv}).
\end{proof}
\end{theorem}

\begin{theorem}[Tractable alpha divergence of two PCs]
\label{thm:trac-alpha-div}
Let $\p$ and $\q$ be compatible PCs over variables $\X$. Then their Rényi's $\alpha$-divergence 
can be exactly computed in $\bigO(\abs{\p}^{\alpha}\abs{\q})$ time for $\alpha\in\N, \alpha>1$ if $\q$ is deterministic or in $\bigO(\abs{\p}\abs{\q})$ for $\alpha\in\R,\alpha\neq 1$ if $\p$ and $\q$ are both deterministic. 
\end{theorem}
\begin{proof}
The proof easily follows from first computing the power circuit of $\p$ and $\q$ according to \cref{thm:real-pow-det} or \cref{thm:trac-natural-pow} in polytime. 
Depending on the value of $\alpha$, the resulting circuits  will have size $\bigO(\abs{\p}^{\alpha})$ and $\bigO(\abs{\q})$ for $\alpha\in\N$ or $\bigO(\abs{\p})$ and $\bigO(\abs{\q})$ for $\alpha\in\R$ and will be compatible with the input circuits. 
Then, since they are compatible between themselves, their product can be done in polytime (\cref{thm:prod-pcs}) and it is going to be a smooth and decomposable PC of size $\bigO(\abs{\p}^{\alpha}\abs{\q})$ (for $\alpha\in\N$) or $\bigO(\abs{\p}\abs{\q})$ (for $\alpha\in\R$), for which the partition function can be computed in time linear in its size.
\end{proof}

\subsubsection{Itakura-Saito Divergence}

\begin{definition}[Itakura-Saito divergence]
\label{def:is-div}
The Itakura-Saito divergence of two PCs $\p$ and $\q$ is defined as
\begin{align}
    \label{eq:is-div}
    &\mathbb{D}_{\mathsf{IS}}(p\parallel q) = \int_{\supp(\p)\cap\supp(\q)} \left ( \frac{\p(\x)}{\q(\x)} - \log \frac{\p(\x)}{\q(\x)} - 1 \right )\: d\X.
\end{align}
\end{definition}

\begin{theorem}[Hardness of Itakura-Saito divergence]
\label{thm:hardness-is-div}
Let $p$ and $q$ be two compatible PCs over variables $\X$. Then computing their Itakura-Saito divergence 
is \#P-Hard.
\end{theorem}
\begin{proof}
We show hardness for the case of binary variables $\X=\{X_1,\ldots,X_n\}$.
Suppose $q$ is an omni-compatible circuit representing the constant function 1, which can be constructed as in \cref{appx:unif-pc}.
As such, integration in \cref{eq:is-div} becomes the summation $\sum_{\val(\X)} {\p(\x)} - \sum_{\val(\X)}\log {\p(\x)}  - 2^{n}$.
Hence, computing $\mathbb{D}_{\mathsf{IS}}$ must be as hard as computing $\sum_{\val(\X)}\log {\p(\x)}$, since the first sum can be efficiently computed as $\p$ must be smooth and decomposable by assumption and the last one is a constant.
That is, we reduced the problem of computing the logarithm of the non-deterministic circuit  (\textsf{LOGPC}, \cref{thm:hardness-log}) to computing $\mathbb{D}_{\mathsf{IS}}$.
\end{proof}

\begin{theorem}[Tractable Itakura-Saito divergence of two circuits]
\label{thm:trac-is-div}Let $p$ and $q$ be two deterministic and compatible PCs over variables $\X$
and with bounded intersectional support ${\supp(\p)\cap\supp(\q)}$, 
then their Itakura-Saito divergence (\cref{def:is-div})
can be exactly computed in time and space $\bigO(\abs{p}\abs{q})$.
\end{theorem}
\begin{proof}
The proof easily follows from noting that the integral decomposes into three integrals over the inner sum: $\int_{\supp(\p)\cap\supp(\q)} \frac{\p(\x)}{\q(\x)}\: d\X$ $- \int_{\supp(\p)\cap\supp(\q)}\log \frac{\p(\x)}{\q(\x)}\: d\X$ - $\int_{\supp(\p)\cap\supp(\q)}1\: d\X.$.
Then, the first integral over the quotient can be solved $\bigO(\abs{p}\abs{q})$ (\cref{thm:ratio-pcs});   the second integral over the log of a quotient of two PCs can be computed in time and space $\bigO(\abs{p}\abs{q})$  (\cref{thm:ratio-pcs},\cref{thm:log-det}) and finally the last one integrates to the dimensionality of  $\abs{{\supp(\p)\cap\supp(\q)}}$, which we assume to exist. 
\end{proof}

\subsubsection{Cauchy-Schwarz Divergence}

\begin{definition}[Cauchy-Schwarz divergence]
\label{def:cs-div}
The Cauchy-Schwarz divergence of two PCs $\p$ and $\q$ is defined as
\begin{align*}
    &\mathbb{D}_{\mathsf{CS}}(p\parallel q) = -\log\frac{\int_{\x\in\val({\X})} p(\x)q(\x)\:d\X}{\sqrt{\int_{\x\in\val({\X})}p^{2}(x)\:d\X\int_{\x\in\val({\X})}q^{2}(x)\:d\X}}.
\end{align*}
\end{definition}

\begin{theorem}[Hardness of Cauchy-Schwarz divergence]
\label{thm:hardness-cs-div}
Let $p$ and $q$ be two structured-decomposable PCs over variables $\X$, then computing their Cauchy-Schwarz divergence (\cref{def:cs-div})
is \#P-Hard.
\begin{proof}
The proof follows by noting that (1) if  $p$ and $q$ are structured-decomposable, then computing the  denominator inside the log can be exactly done in $\abs{\p}^2+\abs{\q}^2$ because they are natural powers of structured-decomposable circuits (\cref{thm:trac-natural-pow}) and hence (2) $\mathbb{D}_{\mathsf{CS}}$ must be as hard as a the product of two non-compatible circuits.
Therefore we can reduce \textsf{MULPC} (\cref{thm:prod-dec}) to computing $\mathbb{D}_{\mathsf{CS}}$.
\end{proof}
\end{theorem}

\begin{theorem}[Tractable Cauchy-Schwarz divergence]
\label{thm:trac-cs-div}Let $p$ and $q$ be two structured-decomposable and compatible PCs over variables $\X$, then their Cauchy-Schwarz divergence (\cref{def:cs-div})
can be exactly computed in time and space $\bigO(\abs{p}\abs{q}\!+\!\abs{\p}^2\!+\!\abs{\q}^2)$.
\begin{proof}
The proof easily follows from noting that the numerator inside the log can be computed in $\bigO(\abs{p}\abs{q})$ time and space as a product of two compatible circuits (\cref{thm:prod-pcs}); and the integrals inside the square root at the denominator can both be solved in $\bigO(\abs{\p}^2)$ and $\bigO(\abs{\q}^2)$ respectively as natural powers of structured-decomposable circuits (\cref{thm:trac-natural-pow}).
\end{proof}
\end{theorem}

\subsubsection{Squared Loss Divergence}

\begin{definition}[Squared Loss divergence]
\label{def:sl-div}
The Squared Loss divergence of two PCs $\p$ and $\q$ is defined as
\begin{align*}
    \label{eq:sl-div}
    &\mathbb{D}_{\mathsf{SL}}(p\parallel q) = \int_{\val(\X)} \left ( \p(\x) - \q(\x) \right )^2\:d\X.
\end{align*}
\end{definition}

\begin{theorem}[Hardness of squared loss]
\label{thm:hardness-sl-div}Let $p$ and $q$ be two structured-decomposable PCs over variables $\X$, then computing their squared loss (\cref{def:sl-div})
is \#P-Hard.
\end{theorem}
\begin{proof}
Proof follows by noting that  the integral decomposes over the expanded square as  $\int_{\val(\X)}\p^2(\x)\:d\X + \int_{\val(\X)}\q^2(\x)\:d\X -2 \int_{\val(\X)}\p(\x)\q(\x)\:d\X$  and that the first two terms can be computed in polytime as natural powers of structured-decomposable circuits  (\cref{thm:trac-natural-pow}), hence computing $\mathbb{D}_{\mathsf{SL}}$ must be as hard as computing the product of two non-compatible circuits.
Therefore we can reduce \textsf{MULPC} (\cref{thm:prod-dec}) to computing $\mathbb{D}_{\mathsf{SL}}$.
\end{proof}

\begin{theorem}[Tractable squared loss]
\label{thm:trac-sl-div}Let $p$ and $q$ be two structured-decomposable and compatible PCs over variables $\X$, then their squared loss (\cref{def:sl-div}
can be exactly computed in time $\bigO(\abs{p}\abs{q}\!+\!\abs{\p}^2\!+\!\abs{\q}^2)$.
\end{theorem}
\begin{proof}
Proof follows by noting that the integral decomposes over the expanded square as  $\int_{\val(\X)}\p^2(\x)\:d\X + \int_{\val(\X)}\q^2(\x)\:d\X -2 \int_{\val(\X)}\p(\x)\q(\x)\:d\X$ and as such each integral can be computed by leveraging the tractable power of structured-decomposable circuits  (\cref{thm:trac-natural-pow}) and the tractable product of compatible circuits (\cref{thm:prod-pcs}) and therefore the overall complexity is given by the maximum of the three.
\end{proof}

\section{Expectation-based queries}

\subsection{Moments of a distribution}
\label{sec:moment-distrib}

\vspace{1em}

\begin{proposition}[Tractable moments of a PC]
Let $\p(\X)$ be a smooth and decomposable PC over variables $\X=\{X_1,\ldots,X_d\}$, then for a set of natural numbers $\mathbf{k}=(k_1,\ldots,k_d)$, its $\mathbf{k}-$moment, defined as 
\label{thm:mom}
\begin{equation*}
\int_{\val(\X)}x_{1}^{k_{1}}x_{2}^{k_{2}}\ldots x_{d}^{k_{d}}\p(\x)\:d\X
\label{eq:mom}
\end{equation*}
can be computed exactly in time $\bigO(\abs{\p})$
\begin{proof}
The proof directly follows from representing $x_{1}^{k_{1}}x_{2}^{k_{2}}\ldots x_{d}^{k_{d}}$ as an omni-compatible circuit comprising a single product unit over $d$ input units, each encoding $x_{i}^{k_i}$, and then applying \cref{cor:omni-prod}. 
\end{proof}
\end{proposition}

\subsection{Probability of logical formulas}
\label{sec:prob-log}

\vspace{1em}

\begin{proposition}[Tractable probability of a logical formula]
\label{prop:log-form}
Let $\p$ be a smooth and decomposable PC over variables $\X$ and $f$ an indicator function that represents a logical formula over $\X$ that can be compiled into a circuit compatible with $\p$.\footnote{For instance by compiling it into an SDD~\citep{darwiche2011sdd,choi2013compiling} whose vtree encodes the hierarchical scope partitioning of $\p$.} 
Then computing $\mathbb{P}_{\p}\left[f\right]$ can be done in $\bigO(\abs{\p}\abs{f})$ time and space.
\begin{proof}
It follows directly from \cref{thm:prod-dec}, by noting that $\mathbb{P}_{\p}\left[f\right]=\mathbb{E}_{\x\sim\p(\X)}\left[f(\x)\right]$ and hence a tractable product between $p$ and $f$ suffices.
\end{proof}
\end{proposition}

\subsection{Expected predictions}
\label{sec:reg-to-circuit}

\vspace{1em}

\begin{figure}
    \centering
    \includegraphics[width=.3\columnwidth]{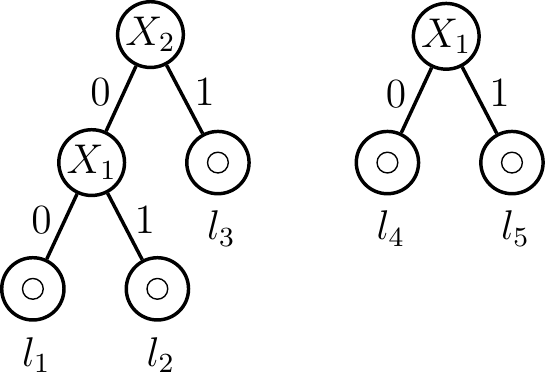}\\[15pt]
    \includegraphics[width=.89\columnwidth]{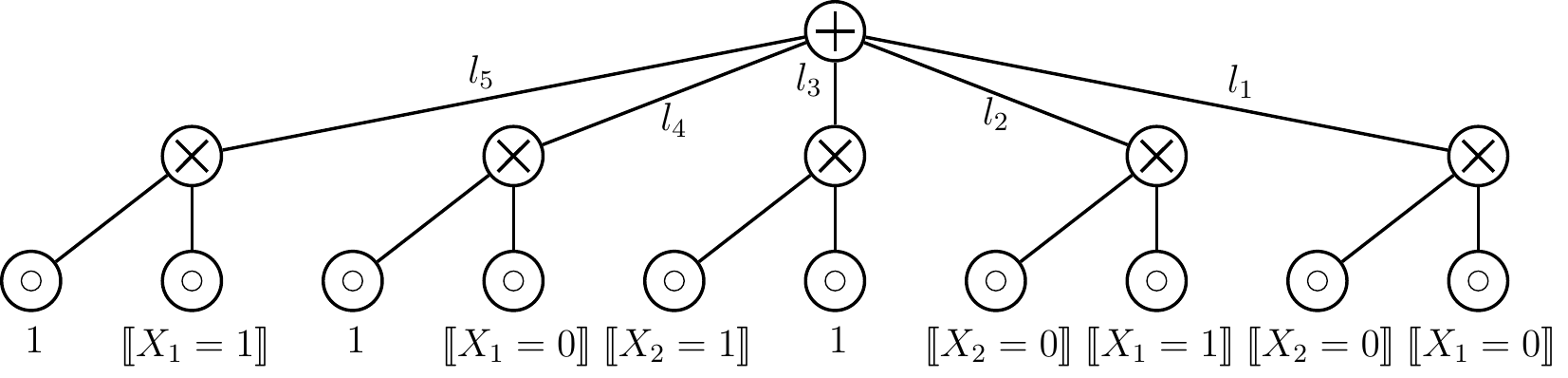}
    \caption{Encoding an additive ensemble of two trees over $\X=\{X_1,X_2\}$ (above) in an omni-compatible circuit over $\X$ (below).
    }
    \label{fig:tree-circuits}
\end{figure}

\begin{example}[Decision trees as circuits]
\label{ex:decision-tree}
Let $\mathcal{F}$ be an additive ensemble of (decision or regression) trees over variables $\X$, also called a \textit{forest}, and computing 
\begin{equation*}
\label{eq:forest}
    \mathcal{F}(\x) = \sum_{\mathcal{T}_{i}\in\mathcal{F}} \theta_{i}\mathcal{T}_{i}(\x)
\end{equation*}
for some input configuration $\x\in\val(\X)$ and each $\mathcal{T}_{i}$ realizing a tree, \ie a function of the form
\begin{equation*}
\label{eq:tree}
    \mathcal{T}(\x) = \sum_{p_{j}\in\mathsf{paths}(\mathcal{T})} l_j \cdot\prod_{X_k\in\scope(p_j)} \id{x_k \leq \delta_k}
\end{equation*}
where the outer sum ranges over all possible paths in tree $\mathcal{T}$, $l_j\in\R$ is the label (class or predicted real) associated to the leaf of that path, and the product is over indicator functions encoding the decision to take one branch of the tree in path $p_j$ if $x_k$, the observed value for variable $X_k$ appearing in the decision node, \ie satisfies the condition $\id{x_k \leq \delta_k}$ for a certain threshold $\delta_k\in\R$. 

Then, it is easy to transform $\mathcal{F}$ into an omni-compatible circuit $p(\X)$ of the form
\begin{equation*}
    \p(\x) = 
    \sum_{\mathcal{T}_{i}\in\mathcal{F}, p_{j}\in\mathsf{paths}(\mathcal{T_i})} l_j \cdot\prod_{X_k\in\scope(p_j)} \id{x_k \leq \delta_k}\cdot \prod_{X_k'\not\in\scope(p_j)}1
\end{equation*}
with a single sum unit realizing the outer sum and as many input product units as paths in the forest, each of which realizing a fully-factorized model over $\X$, and weighted by $l_j$. One example is shown in \cref{fig:tree-circuits}.
\end{example}

\begin{proposition}[Tractable expected predictions of additive ensembles of trees]
\label{prop:exp-pred-dtrees}
Let $\p$ be a smooth and decomposable PC and $f$ an additive ensemble of $k$ decision trees over variables $\X$ and bounded depth. 
Then, its expected predictions can be exactly computed in $\bigO(k\abs{\p})$.
\begin{proof}
Recall that an additive ensemble of decision trees can be encoded as an omni-compatible circuit. Then, proof follows from \cref{cor:omni-prod}.
\end{proof}
\end{proposition}

\begin{algorithm}[!t]
   \caption{\textsc{RGCtoCircuit}($r, \mathsf{cache}_r, \mathsf{cache}_s$)}
   \label{alg:rgc-to-circuit}
   \begin{algorithmic}[1]
   \STATE {\bfseries Input:} a regression circuit $r$ over variables $\X$ and two caches for memoization (i.e., $\mathsf{cache}_r$ and $\mathsf{cache}_s$).
   \STATE {\bfseries Output:} its representation as a circuit $p(\X)$.
   \IF {$r \in \mathsf{cache}_r$}
   \STATE \textbf{return} $\mathsf{cache}_r (r)$
   \ENDIF
   \IF{$r$ is an input gate}
   \STATE $p\leftarrow\inputUnit(0, \scope(r))$
   \ELSIF{$r$ is a sum gate}
   \STATE $\node\leftarrow\{\}$
   \FOR{$i=1$ \textbf{to} $|\ch(r)|$}
   \STATE $\node\leftarrow \node\cup\{\suppCirc(r_i, \mathsf{cache}_s)\}$
   \STATE $\node\leftarrow \node\cup\{\text{\textsc{RGCtoCircuit}}(r_i, \mathsf{cache}_r)\}$
   \ENDFOR
   \STATE $\p\leftarrow\sumUnit(\node, \{\theta_i, 1_1, \ldots, 1_{\ch(\p)}\}_{i=1}^{|\ch(r)|})$
   \ELSIF{$r$ is a product gate}
   \FOR{$i=1$ \textbf{to} $|\ch(r)|$}
     \STATE $\p\leftarrow \prodUnit(\{\text{\textsc{RGCtoCircuit}}(r_i, \mathsf{cache}_r)\}\cup\{\suppCirc(r_j, \mathsf{cache}_s)\}_{j\neq i})$
   \ENDFOR
   \ENDIF
   \STATE $\mathsf{cache}_r(r)\leftarrow \p$
   \STATE \textbf{return} $\p$
\end{algorithmic}
\end{algorithm}

\begin{proposition}[Tractable expected predictions of deep regressors (regression circuits)]
\label{prop:exp-pred-dreg}
Let $\p$ be a structured-decomposable PC over variables $\X$ and $f$ be a regression circuit \citep{khosravi2019tractable} compatible with $\p$ over $\X$, and defined as
\begin{equation*}
f_n(\x) =
    \begin{cases}
    0 &\text{if $n$ is an input } \\
    f_{n_\mathsf{L}}(\x_{\mathsf{L}}) + f_{n_\mathsf{R}}(\x_{\mathsf{R}}) &\text{if $n$ is an AND } \\
    \sum_{c\in\ch(n)} s_c(\x)\left(\phi_{c} +  f_c(\x)\right) &\text{if $n$ is an OR}
\end{cases}
\label{eq:reg-circuit}
\end{equation*}
where $s_c(\x)=\id{\x\in\supp(c)}$. 
Then, its expected predictions can be exactly computed in $\bigO(\abs{p}\abs{h})$ time and space, where $h$ is its circuit representation as computed by \cref{alg:rgc-to-circuit}.
\begin{proof}
Proof follows from noting that \cref{alg:rgc-to-circuit} outputs a polysize circuit representation $h$ in polytime.
Then, computing $\mathbb{E}_{\x\sim\p(\X)}\left[h(\x)\right]$ can be done in $\bigO(\abs{p}\abs{h})$ time and space by \cref{thm:prod-pcs}.
\end{proof}
\end{proposition}

\end{document}